%% file: main.tex
\documentclass[twoside,11pt]{article}

\usepackage{jmlr2e}

\usepackage[dvipsnames,table]{xcolor}
\usepackage{times}
\usepackage{epsfig}
\usepackage{graphicx}
\usepackage{array,multirow}
\usepackage{amsmath}
\usepackage{amssymb}
\usepackage{booktabs}
\usepackage{algorithmicx}
\usepackage{algpseudocode}
\usepackage{capt-of}
\usepackage{subcaption}
\usepackage{longtable}
\usepackage{dcolumn}
\usepackage{placeins}

\newcommand{\HomOpt}{{\texttt{HomOpt}}}
\usepackage{color}
\usepackage[ruled,vlined,linesnumbered]{algorithm2e}
\usepackage[capitalize]{cleveref}
\crefname{section}{Sec.}{Secs.}
\Crefname{section}{Section}{Sections}
\Crefname{table}{Table}{Tables}
\crefname{table}{Tab.}{Tabs.}

\begin{document}




\title{\HomOpt{}: A Homotopy-Based Hyperparameter Optimization Method}

\author{\name Sophia J. Abraham \email sabraha2@nd.edu \\
       \addr Department of Computer Science and Engineering\\
       University of Notre Dame\\
       Notre Dame, IN 46556
       \AND
       \name Kehelwala D. G. Maduranga \email gmaduranga@tntech.edu \\
       \addr Department of Mathematics\\
       Tennessee Tech University\\
       Cookeville, TN 38505
       \AND
       \name  Jeffery Kinnison \email jkinniso@nd.edu \\
       \addr Department of Computer Science and Engineering\\
       University of Notre Dame\\
       Notre Dame, IN 46556
       \AND
       \name Zachariah Carmichael \email zcarmich@nd.edu\\
       \addr Department of Computer Science and Engineering\\
       University of Notre Dame\\
       Notre Dame, IN 46556
       \AND
       \name Jonathan D. Hauenstein \email hauenstein@nd.edu \\
       \addr Department of Applied and Computational Mathematics and Statistics\\
       University of Notre Dame\\
       Notre Dame, IN 46556
       \AND
       \name Walter J. Scheirer \email walter.scheirer@nd.edu \\
       \addr Department of Computer Science and Engineering\\
       University of Notre Dame\\
       Notre Dame, IN 46556
       }

\editor{TBD}

\maketitle

\begin{abstract}
Machine learning has achieved remarkable success over the past couple of decades, often attributed to a combination of algorithmic innovations and the availability of high-quality data available at scale. However, a third critical component is the fine-tuning of hyperparameters, which plays a pivotal role in achieving optimal model performance. Despite its significance, hyperparameter optimization (HPO) remains a challenging task for several reasons. Many HPO techniques rely on naive search methods or assume that the loss function is smooth and continuous, which may not always be the case. Traditional methods, like grid search and Bayesian optimization, often struggle to quickly adapt and efficiently search the loss landscape. Grid search is computationally expensive, while Bayesian optimization can be slow to prime. 
Since the search space for HPO is frequently high-dimensional and non-convex, it is often 
challenging to efficiently find a global minimum. Moreover, optimal hyperparameters can be sensitive to the specific dataset or task, further complicating the search process. To address these issues, we propose a new hyperparameter optimization method, \HomOpt{}, using a data-driven approach based on a generalized additive model (GAM) surrogate combined with homotopy optimization. This strategy augments established optimization methodologies to boost the performance and effectiveness of any given method with faster convergence to the optimum on continuous, discrete, and categorical domain spaces. We compare the effectiveness of \HomOpt{} applied to multiple optimization techniques (\emph{e.g.}, Random Search, TPE, Bayes, and SMAC) showing improved objective performance on many standardized machine learning benchmarks and challenging open-set recognition tasks. 
\end{abstract}

\input{./01introduction}
\input{./02relatedwork}

\input{./03methods}

\input{./04framework}
\input{./05experiments}
\input{./06conclusion}
\input{./07limitations}

\input{./08acknowledgements.tex}

{\small
\bibliography{references}
}

\appendix

\input{appendix.tex}

\end{document}

%% file: 01introduction.tex
\section{Introduction}

Selecting appropriate hyperparameters for a particular machine learning task is a challenging problem due to the vast search space, non-linear or non-monotonic effects on performance, and the complexity of the optimization landscape. Machine learning models consist of two distinct types of parameters: \textit{elementary parameters}, which are learned during model training, and \textit{hyperparameters}, which are higher-level free parameters that structure and control the training process. Most commonly, hyperparameters are set heuristically by practitioners before training, making the process prone to inconsistencies and biases across different individuals or experiments.

Automated machine learning (AutoML) aims to automate the entire machine learning pipeline, with automatic hyperparameter optimization (HPO) as a key subfield. HPO seeks to find the optimal hyperparameters for a given model to achieve the best possible performance on a specific task while improving reproducibility and fairness. By automating the HPO process, the search for optimal hyperparameters becomes more systematic and standardized, ensuring more consistent results when the same optimization algorithm is applied to the same problem. The performance of a model depends on the algorithm's architecture, the training data, and the chosen hyperparameters. Consequently, model selection is not solely an algorithmic determination, as hyperparameters significantly impact an algorithm's capability to learn. Hyperparameters, which can be real-valued, integer-valued, binary, or categorical, need to be set before training and differ from the elementary parameters learned from the data. Hyperparameter search spaces serve as proxy domains for loss functions, which can be defined over nonlinear, non-convex spaces with many oscillations. This complexity makes the optimization process non-trivial. Identifying the best model for a particular learning task involves selecting hyperparameters to achieve the best performance on a specified task. This process is known as the hyperparameter optimization problem.

Various kinds of automated hyperparameter search approaches have been proposed to solve this optimization problem, ranging from simple methods like grid search \citep{Duan_2005} and random search \citep{JMLR:v13:bergstra12a}, to more rigorous methods like Bayesian optimization \citep{NIPS2011_86e8f7ab,Bergstra_2015}, gradient-based learning \citep{Bengio_gradopt_2000, maclaurin2015gradientbased}, and surrogate model approaches \citep{zhang_wang_ji_2015}. These methods have been used in many fields and have their own strengths and drawbacks. For example, grid search and random search are relatively simple to implement but can be computationally expensive and inefficient in exploring the hyperparameter space. Bayesian optimization is more efficient in searching the space but can be sensitive to the choice of acquisition function and prior distributions. Gradient-based learning requires differentiable hyperparameters, which may not always be available, and surrogate model approaches depend on the quality of the surrogate model to guide the search effectively.

Our main contribution in this paper is a Generalized Additive Model (GAM) Surrogate Homotopy Hyperparameter Optimization strategy: \HomOpt{}.  
Following a data-driven approach, GAMs are used to create a sequence of surrogate models
of the hyperparameter space as more data is obtained
and avoids the
``curse of dimensionality'' \cite[Chap.~7]{HandbookStats}.  
Although this creates a discrete sequence of GAMs, it is natural to consider 
how one GAM can be continuously deformed to the next in this sequence. 
A continuous deformation between objects is
called a {\em homotopy} where the homotopies of interest
are those between GAMs.  Therefore, homotopy-based optimization 
techniques are then applicable for hyperparameter optimization
by solving the family of optimization problems 
where the objective function is a homotopy between GAMs.  
For example, homotopy-based optimization methods have been effectively used for solving nonlinear equations and nonlinear optimization problems \citep{bates2013numerically,
griffin2015real,hao2019homotopytraining}. 
In particular, \HomOpt{} employs a homotopy hyperparameter optimization strategy 
as one moves through the sequence of GAMs aiming to boost performance and effectiveness. 
To the best of our knowledge, homotopy-based methods have not been applied within the hyperparameter optimization setting on black-box functions.  

One key aspect is that \HomOpt{} can be used to augment any base optimization strategy.  
For example, as a Bayesian optimization approach is collecting data regarding the hyperparameter space,
one can simultaneously be employing \HomOpt{} on the known data aiming to converge faster.
This is demonstrated in our experiments that compare base methods with ones augmented with
\HomOpt{} on both closed-set and open-set learning. 
While closed-set learning only works on identifying predefined classes, open-set learning~\citep{Scheirer_2013_TPAMI} trains models that have incomplete knowledge of the world they must operate in and allow the incremental learning setting, where newly identified classes are added to the recognition model over time.  We tune both a multitude of common machine learning-based classifiers and the Extreme Value Machine~(EVM) \citep{rudd2017extreme}, which is a scalable nonlinear open-set classifier. 

To summarize, our contributions are the following: 
\begin{enumerate}
    \item We demonstrate that \HomOpt{}, which
    does not rely on strong assumptions about the behavior of the underlying objective function,
    augments traditional methods and converges to optima faster than multiple base methods alone by approximating regions of interest rather than the entire hyperparameter surface.
    \item We demonstrate the use of \HomOpt{} across continuous, discrete, and categorical search domains, showcasing its versatility and effectiveness in various settings.
    \item We release an easy to use, flexible, and open-sourced software package to apply \HomOpt{} to model search for rapid deployment in general search problems.
\end{enumerate}

%% file: 02relatedwork.tex
\section{Related Work}
\noindent \textbf{Hyperparameter Optimization.} Hyperparameter optimization methods involve searching for optimal hyperparameter values to effectively identify high-performing models within the hyperparameter space. Non-Bayesian approaches, such as hand-tuning and grid search~\citep{Duan_2005}, are simple to use.  However, they rely upon adequate domain knowledge, which may not readily be available. Furthermore, these methods may overlook optimal values in continuous domains and inevitably prove brittle when applied to unseen cases~\citep{li2017hyperband}. 
Random search~\citep{JMLR:v13:bergstra12a} mitigates this issue by removing the requirement of discretizing the search space and provides a larger coverage of the hyperparameter space. Although random search is simple to use, it is often inefficient sampling-wise. In order to narrow the scope of the search, multiple software frameworks for hyperparameter search based on random search have been proposed,
including those by
\cite{Bergstra_2015, NIPS2011_86e8f7ab, betro_1992, WU201926,Bengio_gradopt_2000, maclaurin2015gradientbased, ilievski_akhtar_feng_annette_2017}. 

\noindent \textbf{Population-Based Approaches.} Population-based algorithms take inspiration from biology and improve upon computational efficiency over purely random search-based methods~\citep{loshchilov2016cma}. These methods include Evolutionary Algorithms and swarm algorithms like particle swarm optimization (PSO)~\citep{Boeringer_2005}, which iteratively update the generation of hyperparameters with a stochastic velocity term. While effective for lower dimensional spaces, methods like PSO can get stuck in a local optimum for high dimensional, complex scenarios with low convergence rates over the iterative process~\citep{kennedy1995particle}. 


\noindent \textbf{Bayesian Optimization.} 
In order to perform the search 
using statistical analysis, Bayesian optimization methods have been proposed~\citep{Bergstra_2015, NIPS2011_86e8f7ab, betro_1992, WU201926} based on Bayes' theorem. It sets a prior over the optimization function and gathers the information from the previous sample to update the posterior of the optimization function. A utility function selects the next sample point to maximize or minimize the optimization function. One example of a popular Bayesian method is the Sequential Model-Based Algorithm Configuration (SMAC)~\citep{smac3_documentation} consisting of Bayesian optimization combined with a simple racing mechanism on the instances to efficiently decide which of two configurations performs better. 

Tree-Structured Parzen Estimators (TPE)~\citep{NIPS2011_86e8f7ab, Bergstra_2015}, another Bayesian method, is a sequential model-based optimization (SMBO) approach. SMBO methods sequentially construct models to approximate the performance of hyperparameters based on historical measurements, and then subsequently choose new hyperparameters to test with based on a constructed model. 

\noindent \textbf{Gradient-based Approaches.} 
Gradient-based optimization methods~\citep{Bengio_gradopt_2000, maclaurin2015gradientbased} compute gradients of cross-validation performance with respect to all hyperparameters by chaining derivatives backwards through the entire training procedure. This is advantageous over other methods since information regarding the shape of the objective surface and behaviors including extrema in the parameter space can be acquired. Hyperparameter gradients are computed by reversing the dynamics of stochastic gradient descent. Gradients enable the optimization of the hyperparameters, including step-size, momentum schedules, weight initialization distributions, richly parameterized regularization schemes, and neural network architectures. However, information about the gradients is often unavailable, computing gradients is computationally expensive, and 
gradient-based approaches suffer from inefficiency when learning long-term dependencies~\citep{Bengio_94_279181}.

\noindent \textbf{Surrogate-based Approaches.}  Surrogate-based optimization methods~\citep{Eggensperger_1970, Weicheng_XIE2021107701, mcleod2018optimization} are used when an objective function is expensive to evaluate. The Surrogate Benchmarks for Hyperparameter Optimization~\citep{Eggensperger_1970} uses the following strategy: cheap-to-evaluate surrogates of real hyperparameter optimization benchmarks that yield the same hyperparameter spaces and feature-similar response surfaces. Specifically, this approach trains regression models on data representing a machine learning algorithm’s performance under a broad range of hyperparameter configurations and then cheaply evaluates hyperparameter optimization methods using the model’s performance predictions instead of the actual algorithm. In~\cite{mcleod2018optimization}, a Gaussian Process-based (GP) model was used to identify a convex region and a probability-based approach was used to estimate a convex region centered around the posterior minimum. Our approach uses the exploitation from multiple optimization techniques to identify the region of interest for surrogate approximation.

\noindent \textbf{Homotopy-based Approaches.}
Continuation or homotopy methods \citep{RHEINBOLDT1981103,AllgowerGeorg} have long served as a useful technique in numerical methods and result in accurate classification performance without excessive computational effort. Preliminary results indicate the potential of this approach with respect to both training time and classification accuracy and it is also known as globally convergent and solution exhaustive for nonlinear polynomials~\citep{AllgowerGeorg,bates2013numerically}. There has also been recent interest in homotopy continuation for training and analyzing neural networks. This was first presented by \citep{Chow91} and thereafter by \citep{Pathak2018ParameterCW,hao2019homotopytraining,Mehta2022LossSurface}. To train and initialize neural networks, data continuation and model continuation are used with similar objectives of decomposing the original task into a sequence of tasks with increasing levels of difficulty. To the best of our knowledge, the continuation method for training a neural network is used mainly for elementary parameter optimization, and is also data specific. We are extending the idea of using homotopy optimization into general hyperparameter optimization.

%% file: 03methods.tex
\section{Homotopy-Based Hyperparameter Optimization}

We propose a new data-driven hyperparameter optimization approach
based on the following 
two core aspects.
First, surrogates are used to model the objective
since the function and its gradient are computationally expensive to evaluate.  In particular, a sequence of surrogate models
is constructed as new data is gathered.  
A continuous deformation from the current surrogate model
to the updated one that includes the new data is formed.
Second, our approach uses this continuous deformation
to construct a homotopy path between a local minimum for the current
model and a local minimum for the updated model.
Since the data may be generated by any base optimization strategy,
our approach can be used to augment other strategies
aiming to converge in fewer iterations.

The first step is to select a family of surrogate models.  
Some examples include polynomial spline fitting with several degrees of freedom, radial basis function interpolation, and a generalized additive model (GAM)~\citep{hastie1990generalized},
which is a type of statistical model that is used to describe the relationship between a response variable and one or more predictor variables. 
GAMs are similar to generalized linear models (GLMs), but they allow for nonlinear relationships between the 
response (output or target variable we try to predict) and predictor variables (input features used to make the prediction) by using smooth functions to model the relationships. These can be estimated using penalized regression techniques, such as penalized splines, and they can provide more flexible and accurate models than GLMs in many cases. 
The results of some initial empirical experiments
along with some theoretical advantages described below
suggest that the extrapolation behavior is more reasonable for 
GAMs and, for multidimensional data, 
is more suitable and provides a more accurate fit.
In our implementation we utilize GAMs
via PyGAM~\citep{daniel_serven_2018_1476122}.

From a theoretical standpoint, GAMs provide a more interpretable model than other surrogate models as the smoothing functions that are used to model the relationships between the response and predictor variables can be visualized and analyzed directly. 
This can be especially useful for understanding and explaining the underlying patterns and relationships in the data, which can be difficult to do with more complex and opaque models like random forests. Additionally, GAMs can provide more accurate predictions for certain types of data, such as data with nonlinear or non-monotonic relationships between the response and predictor variables.
Moreover, relationships between the response and predictor variables in GAMs can provide insight into the importance of each feature.  For example, the magnitude and significance of the coefficients of the smoothing functions 
can be used to determine the relative importance of each 
feature, and the shape of the smoothing function 
can provide additional information about the nature of the relationship between the response and predictor variables. 
Although this is out of the scope of this paper,
it could be a useful tool for understanding and interpreting the results of the hyperparameter
optimization process.

\begin{figure}
\begin{center}
\centering
    \includegraphics[width=0.75\textwidth]{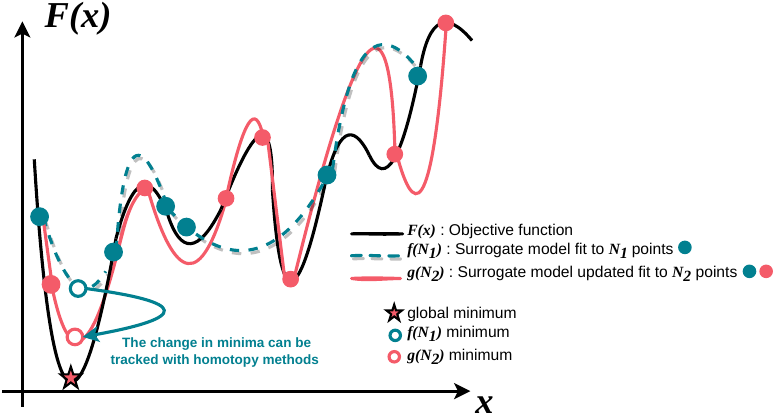}
T:    \caption{Illustration of homotopy parametrization between the initial samples and updated data samples. The black line represents the objective function that is being evaluated. The first surrogate model \textbf{$f(N_1)$} is fit on the initial set of samples indicated by the blue circles. With more samples (indicated in pink), the updated surrogate $g(N_2)$ yields a new minimum. As the number of data samples increases, the approximation of the minimum improves. These changing minima can be tracked with homotopy methods.}
    \label{fig:homotopy}
\end{center}
\end{figure}

Following the selection of the family of surrogate models,
the second step is to construct a continuous deformation
between surrogate models as new data is collected
and then utilize homotopy methods to track a local minimum 
along this deformation as illustrated in Figure~\ref{fig:homotopy}.
For a general overview of homotopy methods,
with a focus on nonlinear polynomial functions,
see \citep{bates2013numerically}.
In the context of optimization using surrogate
models that depend upon continuous variables, 
local minima are critical points of
the surrogate model, \textit{i.e.}, the gradient 
of the surrogate model vanishes at each local minimum.
Although computing all critical points of the objective
function using homotopy methods has shown
to be useful in some applications \citep{Mehta2022LossSurface,
Baskar2022Saddlegraphs}, the approach utilized
here does not rely upon computing all critical
points to provide a scalable framework to higher-dimensional
and non-polynomial systems.


In our approach, one starts with a GAM $f(x)$
fit to $N_1$ points.  Viewing $f$ as an objective function of an unconstrained
optimization problem, suppose that one has computed a local minimum $x_{\text{old}}$
for~$f$.  As new data is collected, one creates a new GAM $g(x)$ fit to $N_2$ points.
Thus, one can construct a continuous family of functions that deforms 
from $f$ to $g$, such as the linear family
\begin{equation}
H(x,t) = t\cdot f(x) + (1-t) \cdot g(x)
\hbox{~~~~~where~~~~~$H(x,1) = f(x)$~~~~~and~~~~~$H(x,0)=g(x)$.}
\end{equation}
Therefore, one now has a family of unconstrained optimization problems
with objective function $H(x,t)$ and a known local minimum when $t=1$
at $x_{\text{old}}$.  This yields a homotopy path of local minima of $H(x,t)$
parameterized by $t$ that emanates from $x_{\text{old}}$ at $t=1$.
When $f$ and $g$ are analytic, one can use standard homotopy theory
\citep{SW05} to provide guarantees on existence and smoothness
of such a path based on $f$ and $g$ via the implicit function theorem
that ends at the desired local minima $x_{\text{new}}$ of $g$ at $t=0$.
However, since we aim to apply this framework to general hyperparameter optimization 
problems, we do not rely upon smoothness or differentiability
by employing techniques which do not require known derivatives.
In particular, a high-level description of the homotopy optimization framework \HomOpt{} is given in Algorithm \ref{algo:pshho}.  The key step of that algorithm
is in line 15 which is described in Algorithm \ref{algo:homotopy}. 
In particular, line 7 of Algorithm \ref{algo:homotopy} uses Nelder-Mead optimization \citep{Nelder65} to minimize the homotopy function. 
Nelder-Mead does not require or use function gradient information and is appropriate for optimization problems where the gradient is unknown or cannot be reasonably computed.

\begin{algorithm}[t]
    \SetAlgoLined
    \textbf{Input} $\mathcal{T}$ time used for optimization, $\mathcal{N_T}$ trials used for optimization, $\mathcal{W}$ samples used for surrogate approximation, $\mathcal{D}$ distance threshold controlling localization of surrogate, \texttt{Inner\_Method} which is the method 
    used to gather new sample data points, $k$ the proportion of trial data to use \\
    
    \textbf{Output} Optimal hyperparameter set \\
    $\mathcal{C_T} \gets 0$ \Comment{Variable to track the number of completed trials} \\
    $trial\_data \gets $ empty array \\
    \While{$\text{TIME} \leq  \mathcal{T} \textup{ or } \mathcal{C_T} \leq \mathcal{N_T}$}{

    \uIf{$\mathcal{C_T} < \mathcal{W}$ \textup{or} $\mathcal{C_T} \ \mathrm{mod}\  5 \in \{0,2\}$}{
    $hparams \gets$ Generate hyperparameters using \texttt{Inner\_Method} \\
}{\uElseIf{$\mathcal{C_T}\ \mathrm{mod}\ 5 \in \{3,4\}$}{
    $svar \gets \mathrm{variance}(\text{best 10 hyperparameter sets in }trial\_data) \cdot \mathcal{D}$ \\
    $hparams \gets$ lowest-loss hyperparameters in $trial\_data$ \\
    $hparams \gets hparams + \mathrm{uniform}(\mu=-svar, \sigma^2= 2\cdot svar)$
}\Else{

    Fit $f$ to the $\mathrm{round} (k \cdot \mathcal{C_T} )$ most recent entries of $trial\_data$ (hyperparameters as features and losses as the targets) \\
    Fit $g$ to $trial\_data$ \\
    Use homotopy optimization (Alg.~\ref{algo:homotopy}) to minimize along $H(x,t) = t\cdot f(x) + (1-t)\cdot g(x)$ with the lowest-loss hyperparameters in $trial\_data$ as the initial guess \\
    Predict loss values for all of the hyperparameters returned by Alg.~\ref{algo:homotopy} using $g$ \\
    $hparams \gets$ lowest-loss hyperparameters from the previous line
    }
    }
    $loss \gets $ Evaluate $hparams$ on the function being optimized \\
    Append $(hparams, loss)$ to $trial\_data$
}
$hparams \gets $ the lowest-loss hyperparameters in $trial\_data$ \\
\textbf{return} $hparams$
     \caption{Homotopy Optimization Framework}
     \label{algo:pshho}
\end{algorithm}


\begin{algorithm}[t] 
\caption{Homotopy Optimization}
\label{algo:homotopy}
\textbf{Input} $\mathcal{N}$ number of steps along interval,
homotopy function $H(x,t)$, $x^{(0)} = x^0$ a local minimum of $H(x,1)$\\ 
\textbf{Output} Local minimum of $H(x,0)$ \\
  {$\Delta$ $\gets$ {$1/\mathcal{N}$}}\\
    $t\gets 1$\\
    \For{$k$ $\gets$ $1$ to $\mathcal{N}$}{             
        $t$ $\gets$ {$t-\Delta$}\\
        Use Nelder-Mead optimization to minimize $H(x,t)$
        starting with $x^{(k-1)}$ to obtain $x^{(k)}$.
    }
    \Return {$x^{(N)}$}

\end{algorithm}
\subsection{One-Dimensional Illustration}

\begin{figure}[ht]
  \subfloat[Start of optimization]{
	\begin{minipage}[1\width]{
	   0.52\textwidth}
	   \centering
	   \includegraphics[width=1\textwidth]{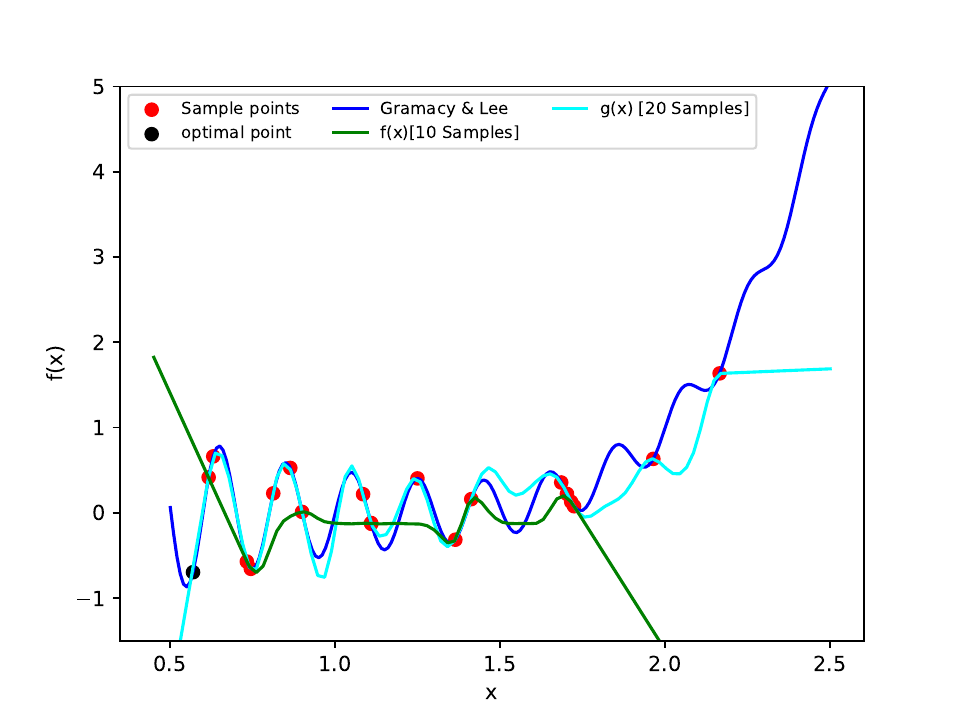}
	\end{minipage}}
 \hfill	
  \subfloat[After 8 optimization steps]{
	\begin{minipage}[1\width]{
	   0.52\textwidth}
	   \centering
	   \includegraphics[width=1\textwidth]{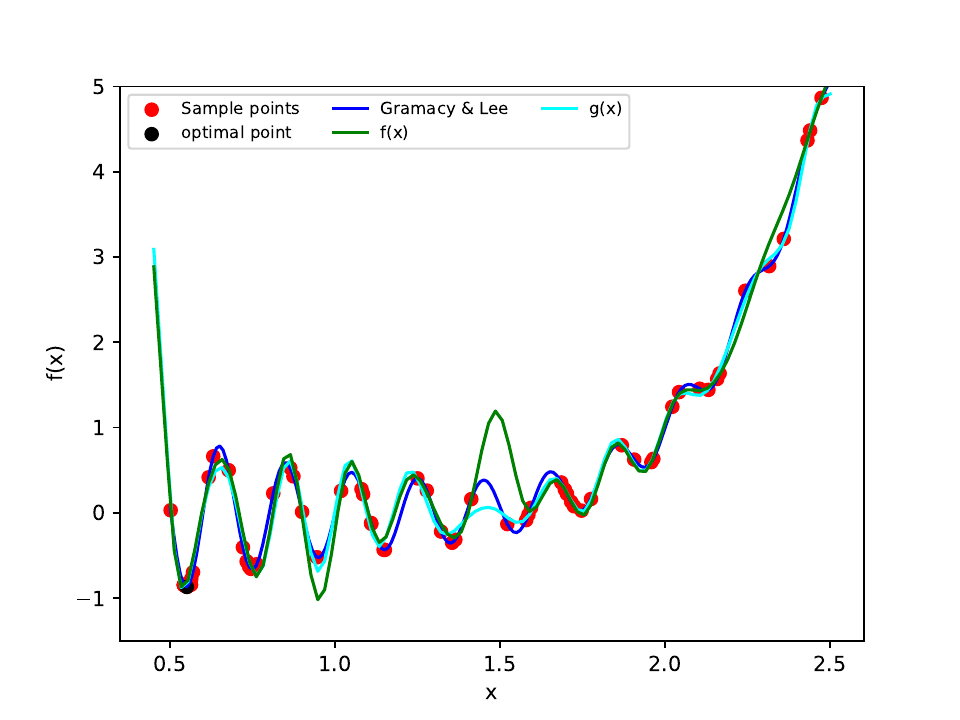}
	\end{minipage}}
 \caption{The Plot of the Gramacy and Lee function on the domain $[0.5,2.5]$ (blue curve). The strategy used by \HomOpt{} is to find a homotopy continuation path from the minimum of $f(x)$ to the minimum of the $g(x)$.
      (a) Local surrogate approximations of $f(x)$ (green curve), and surrogate approximation of $g(x)$ (cyan curve) at the initial stages. (b) Local surrogate approximations of $f(x)$ (green curve), and surrogate approximation of $g(x)$ (cyan curve) after eight homotopy optimization steps.}
     \label{fig:example1D}
\end{figure}

To illustrate \HomOpt{}, we consider the optimization of the test function $p(x)$ defined by \cite{GramacyLee} on the domain $[0.5, 2.5]$, where
\begin{equation}
p(x) = \frac{\sin (10\pi x)}{2x} + (x-1)^4.
\end{equation}
The plot of $p(x)$ is shown in blue in Figure~\ref{fig:example1D}.
Consider two GAMs $f$ and $g$ constructed from $10$ and $20$ sample points, respectively,
as shown in Figure~\ref{fig:example1D}(a). 
Note that the number of samples used in $f$ and $g$ were selected arbitrarily for illustration purposes.
The strategy for \HomOpt{} is to 
consider the homotopy path from a known local minimum of $f(x)$ to a local minimum of $g(x)$
that we want to compute.  
Figure~\ref{fig:example1D}(b) shows the optimized surrogate curves $f(x)$ and $g(x)$, along with the corresponding optimal point obtained after eight iterations of the \HomOpt{} method
demonstrating convergence to the global minimum.


\subsection{Two-Dimensional Example}

\begin{figure}[ht]
   \subfloat[Griewank Function]{
 	\begin{minipage}[c][1\width]{
 	   0.45\textwidth}
 	   \centering
 	   \includegraphics[width=1\textwidth]{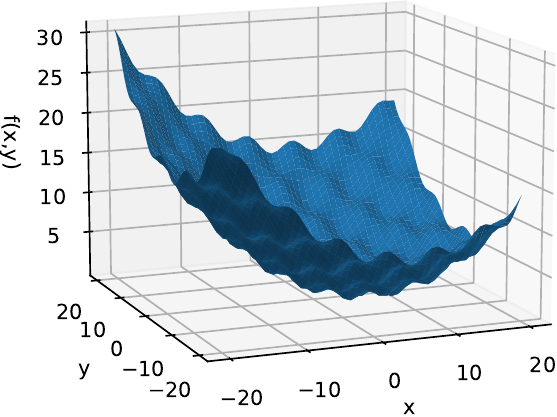}
 	\end{minipage}}
  \hfill 	
   \subfloat[\HomOpt{} convergence to optimal point]{
 	\begin{minipage}[c][1\width]{
 	   0.45\textwidth}
 	   \centering
 	   \includegraphics[width=1\textwidth]{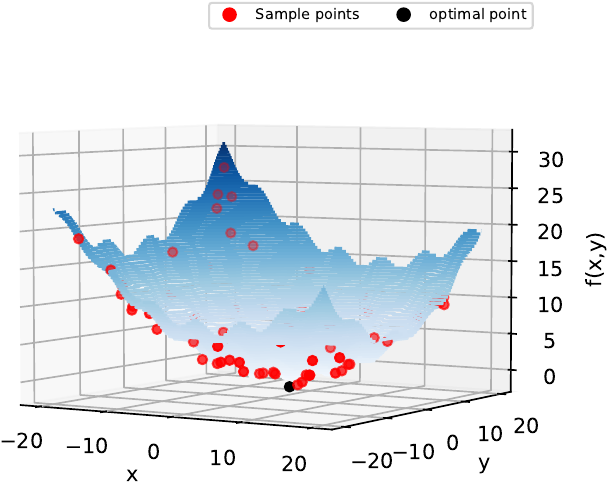}
 	\end{minipage}}
 \caption{(a) Plot of the modified Griewank function on the domain $[-20,20]^2$.
      (b) \HomOpt{} converged towards the optimal point shown (black point) using 100 randomly sampled points (red points).}
      \label{fig:example2D1}
 \end{figure}

 Now consider the two-dimensional case. The Griewank function, introduced by Griewank \citep{Griewank}, is a standard test example in optimization problems in the 2D plane. However, to prevent the global minimum from being located at the origin and increase the difficulty of the problem, a modified function is considered over the domain $[-20, 20]^2$:
 \begin{equation}
     g(x,y) = \frac{(x-5)^2 +(y+3)^2}{40}-\cos(x-5) \cdot \cos\left(\frac{y+3}{\sqrt{2}}\right)+1
 \end{equation}
which is plotted in Figure~\ref{fig:example2D1}(a).

\begin{figure}[ht]
   \subfloat[Surrogate from 10 samples]{
 	\begin{minipage}[c][1\width]{
 	   0.3\textwidth}
 	   \centering
 	   \includegraphics[width=1\textwidth]{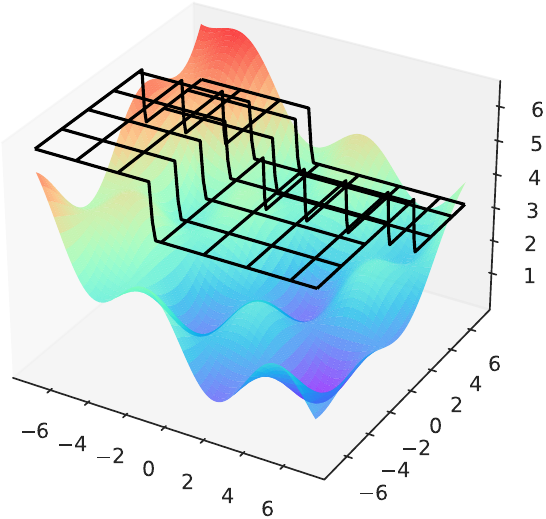}
 	\end{minipage}}
  \hfill 	
   \subfloat[Surrogate from 20 samples]{
 	\begin{minipage}[c][1\width]{
 	   0.3\textwidth}
 	   \centering
 	   \includegraphics[width=1\textwidth]{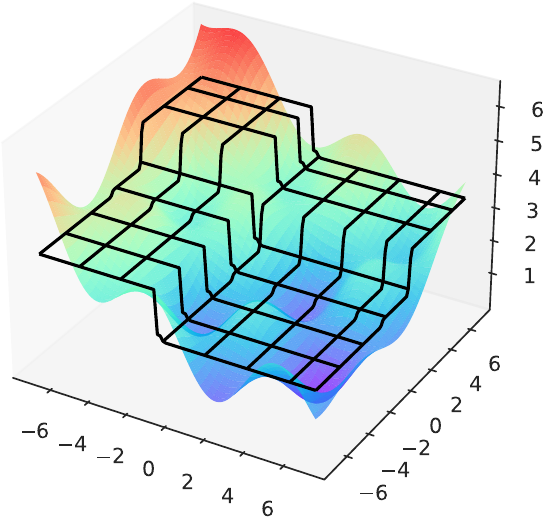}
 	\end{minipage}}
    \hfill 	
   \subfloat[Surrogate from 40 samples]{
 	\begin{minipage}[c][1\width]{
 	   0.3\textwidth}
 	   \centering
 	   \includegraphics[width=1\textwidth]{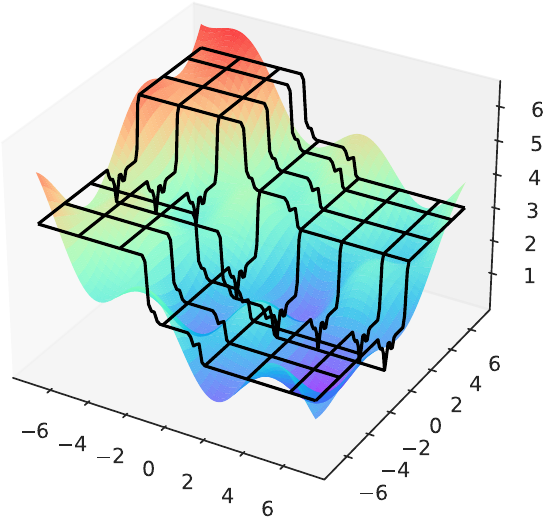}
 	\end{minipage}}
 \caption{Plots of the modified Griewank function on the domain $[-7,7]^2$ with the surrogate approximation plotted as a black grid fit to different numbers of samples. As the numbers of samples increases,  the surrogate function begins to get a better approximation of the underlying function and converge towards the global optimum.}
      \label{fig:example2D_lambda}
 \end{figure}

In Figure \ref{fig:example2D1}(b), the optimal point (in black) for the Griewank function is shown along with the sampled points (in red). \HomOpt{} was able to successfully converge to the optimal point within 100 sampled points. The Griewank function is relatively simple and a low-dimensional problem and as such 100 sampled points was sufficient to accurately approximate the objective function and converge to the optimum. It is important to note that for more complex and higher dimensional problems, it may be necessary to use a larger number of sampled points. As depicted in Figure \ref{fig:example2D_lambda}, the black grid representing the approximation of \HomOpt{} improved as more sample points were added, learning the overall shape of the Griewank function within 40 sample points.

%% file: 04framework.tex
\subsection{Optimization Software Framework} 
The baseline search framework utilized
the massively Scalable Hardware-Aware Distributed Hyperparameter Optimization~(SHADHO) software framework \citep{shadho_8354190}. SHADHO presents a comprehensive survey of existing hyperparameter optimization methods and facilitates baseline Search Strategies, such as Random search, Bayesian optimization, PSO, SMAC, and TPE. 
SHADHO is designed to scale effectively in distributed computing environments, making it suitable for optimizing complex machine learning models with large hyperparameter search spaces. In addition to serving as a hyperparameter optimization tool, SHADHO has the added utility of considering hardware specifications when conducting a search. This framework calculates the relative complexity of each search space and monitors performance on the learning task over all trials. These metrics are then used as heuristics to assign hyperparameters to distributed workers based on their hardware. We extended SHADHO's framework by integrating \HomOpt{}. This extension is now available in the most recent release of SHADHO\footnote{Code can be found at \url{https://github.com/jeffkinnison/shadho}}.

%% file: 05experiments.tex
\section{Experiments}
\label{experiments}


We evaluate \HomOpt{} on a multitude of tasks. We begin with a collection of machine learning benchmarks for classification tasks on tabular data using a Multi-Layer Perceptron (MLP), Support Vector Machine (SVM), Random Forest, XGBoost, and Logistic Regression provided through HPOBench \citep{eggensperger2021hpobench}. 

Additionally, we include a set of difficult \textit{open-set} classification experiments. In the open-set scenario, models have incomplete knowledge of the world they must operate in and unknown classes are queried during testing \citep{Scheirer_2013_TPAMI}. Open-set classification is notoriously sensitive to hyperparameters and provides a complex scenario where the loss landscape is arbitrary, polluted by noise, and may consist of steep gradients resulting in the absence of regularity. This set of experiments also captures how well \HomOpt{} can boost methods in changing environments and unseen conditions. 

We use the Extreme Value Machine (EVM) \citep{rudd2017extreme}, which is a scalable nonlinear classifier that supports open-set classification
by rejecting inputs that are beyond the support of the training set. 
The EVM relies on a strong feature representation and every represented sample in the feature representation becomes a point. It utilizes a binning strategy that groups all the points in their feature representation by their corresponding label. These bins are utilized to create a ``1 vs. rest'' classifier for each known class. It generates a classifier where a Weibull distribution is fit on the data for each known class and is made to avoid the negative data points (unknown classes). This process is repeated for all known classes. When a new data point (a sample represented by its feature vector) is provided to the EVM, it is evaluated in the feature space, and the probability of the point belonging to each representative class is determined.

\begin{table}[htbp]
    \centering
    \small
    \renewcommand{\arraystretch}{1.3}
    \begin{tabular}{p{3cm}p{8cm}p{2cm}}
        \toprule
        \textbf{Parameter} & \textbf{Description} & \textbf{Domain} \\
        \midrule
        Threshold & Probability threshold used to determine if an input coordinate point should be classified as ‘unknown’ if the point falls below this probability of inclusion. & $[0,1]$\\
        Tailsize & Defines how many negative samples are used to estimate the model parameters. & $[0,0.5]$\\
        Cover threshold & The probability threshold used to eliminate extreme vectors if they are covered by other extreme vectors with that probability. & $[0,1]$\\
        Distance multiplier & The multiplier to compute margin distances. & $[0,1]$\\
        Distance function & The distance function used to compute the distance between two samples. & Cosine or Euclidean\\
        \bottomrule
    \end{tabular}
    \caption{Hyperparameters used in training the EVM.}
    \label{tab:model_parameters}
\end{table}

The hyperparameters involved in training the EVM are included in Table \ref{tab:model_parameters}. The datasets used in the HPOBench experiments can be found in Table \ref{table:openML-task-ids} and the benchmark details along with their configuration spaces can be found in Table \ref{tab:hpobench}. For further specifications regarding specifics on the benchmarks and datasets from HPOBench please refer to the original paper \citep{eggensperger2021hpobench}. 

\subsection{Evaluation Criteria}
For the HPOBench classification benchmarks, the optimizer performance is evaluated based on the validation performance (\textit{i.e.}, the objective value seen by the optimizer). This objective value was minimized over $1 - \text{accuracy}$ and a summary across all of the benchmark experiments are reported. Additional results including the corresponding values for test scores are included for each experiment in Appendix \ref{sec:extended-results}.

With respect to the HPOBench experiments, we illustrate the performance of each method by plotting the \textit{simple regret} of the minimization problem computed as 
\begin{equation}
    R_t = f(x_{\text{best-so-far}}) - \text{min}_{i=1}^{t} f(x_i)
\end{equation}
where $R_t$ is the regret at iteration $t$, $\min_{i=1}^{t} f(x_i)$ is the global minimum for that dataset found by taking the minimum value across all benchmarks and methods for that dataset, and $f(x_{\text{best-so-far}})$ is the minimum value found so far. We use a log scale for regret values for easier visualization.

For the open-set experiments, the trained EVM is optimized to minimize the loss on a validation set of data containing samples from both known and unknown classes within the number of search trials listed in Table \ref{tab:table-exp}. This loss is defined as the negative $F1$ score with \textit{weighted} averaging: 
\begin{equation}
    \hbox{$F1 = \dfrac{2\cdot\hbox{Precision}\cdot\hbox{Recall}}{\hbox{Precision}+\hbox{Recall}} = \dfrac{2\cdot \hbox{TP}}{2\cdot \hbox{TP}+\hbox{FP}+\hbox{FN}}$}
\end{equation}
where TP, FP, and FN are the number of true positives, false positives, and false negatives, respectively.

\begin{table}[t]
\renewcommand{\arraystretch}{1.3}
\resizebox{\textwidth}{!}{%
\begin{tabular}{@{}lccccccc@{}}
\toprule
\multicolumn{1}{c}{\textbf{Experiment}} &
  \textbf{\# Search Trials} &
  \textbf{\# Known Classes} &
  \textbf{\# Unknown Classes} &
  \textbf{Training} &
  \textbf{Validation} &
  \textbf{Negatives} &
  \textbf{Testing} \\ \midrule
\multicolumn{1}{c}{\textbf{MNIST}} &
  \multicolumn{1}{c}{1000} &
  \multicolumn{1}{c}{6} &
  \multicolumn{1}{c}{4} &
  \multicolumn{1}{c}{28824} &
  \multicolumn{1}{c}{12000} &
  \multicolumn{1}{c}{19176} &
  10000 \\ 
\multicolumn{1}{c}{\textbf{LFW}} &
  \multicolumn{1}{c}{1000} &
  \multicolumn{1}{c}{34} &
  \multicolumn{1}{c}{5715} &
  \multicolumn{1}{c}{1333} &
  \multicolumn{1}{c}{2481} &
  \multicolumn{1}{c}{6110} &
  3309 \\ 
 \bottomrule
\end{tabular}%
}
\caption{Summary of dataset characteristics for experimental sets.}
\label{tab:table-exp}
\end{table}

\subsection{Experimental Setup}

\HomOpt{} can be used to augment any base HPO strategy. We thus compare the performance boost of \HomOpt{} against the base performance of common popular hyperparameter optimization approaches: Random Search, Bayes, TPE, and SMAC. The numbers of iterations and samples used in the different optimization methods were chosen based on previous empirical experiments and theoretical considerations to balance the trade-off between the computational cost and the expected performance of the method. With TPE we used 20\% of the results for the top-k mixture model seeded with 10 random search iterations and 10 generated candidates samples. In the Bayes examples, we used 20 random search iterations with 10 candidate samples. And for the SMAC runs, we used 20 random search iterations with 20 candidate samples. \HomOpt{} similarly was initialized with 20 evaluations ($\mathcal{W}$ = 20) from each of the base strategies. 

All experiments were run for 500 iterations across 5 separate seeds and averaged. The HPOBench classification experiments were run on a multitude of dataset tasks from OpenML \citep{vanschoren2014openml}, which is open platform for sharing datasets, and the open-set experiments were conducted on the handwritten digits dataset MNIST \citep{deng2012mnist} and Labeled Faced in the Wild (LFW) \citep{LFWTech}. 

For the open-set experiments, we chose MNIST and LFW to consider two datasets of different complexity and modified each classification task  to convert from a closed-set task to an open-set one.  Table~\ref{tab:table-exp} summarizes the dataset experiments in terms of number of search trials, number of known and unknown classes, and number of training, validation, negatives, and testing samples. Negatives are samples from the unknown class without labels used to better inform the ``1 vs. rest'' classifiers when training the EVM. The number of trials for each experiment was determined and adjusted according to the time required to train a single model for the given dataset. If the dataset contained many images per class, fitting the EVM to the feature vectors for large samples required longer compute time and thus the number of search trials was reduced. 

For the handwritten digits dataset MNIST, we designate handwritten digits 0 to 5 as the known classes and digits 6 to 9 as the unknown. Each image is represented as a flattened vector of the image pixels (784 features). For Labeled Faces in the Wild (LFW), we use classes with 30 or more  face image samples to designate the known classes (34) and assign the remaining classes (5715) as the unknown set. The network used to extract features is an ArcFace \citep{Arcface_2018} based feature extractor \citep{albiero2020does} trained on the MS-Celeb-1M dataset \citep{guo2016ms} resulting in a 512 dimensional feature vector. 

Threshold, cover threshold, and distance multiplier are sampled from a uniform distribution of range~$[0, 1]$. The fitting algorithm for the EVM requires that the tailsize not exceed greater than half the number of training samples. For this reason, the tailsize hyperparameter was sampled from a uniform distribution between the range 
$[0, 0.5]$ and used as a multiplier to determine how many negative samples were included in estimating the model parameters. 

Our default values for the parameters in the experiments were values found to be effective in empirical studies on simple problems and common choices found in literature. For all experiments, the distance threshold $\mathcal{D}$ is computed by the variance of the best 10\% of the observed sample points scaled by 0.005 (also known as the jitter strength) as the local perturbation search around the observed minimum in Algorithm \ref{algo:pshho}. In all the experiments we use 5 iterations which are the number of minimizations to compute the homotopy ($\mathcal{N}$). This strikes a balance between the computational costs and finding a good solution.  Fewer iterations may not find a good solution, while more iterations may be computationally expensive without much benefit in terms of improved performance. We use a $k$ of 0.5 which are the fraction of complete trials used to train one of the GAMs. Using half the completed trials provides a good balance between using enough data to train the model and not overfitting to the data. A smaller value of $k$ would mean the GAM is trained on fewer data points, which may result in underfitting. A larger value of $k$ would mean the GAM is trained on more data points, which may result in overfitting.The GAM model surrogates use a penalty term on the smooth functions of $10^{-4}$ and 25 splines for all experiments. The penalty term helps to control the complexity of the surrogate model. A smaller penalty term would result in a model that fits the data well but is likely to overfit. A larger penalty term would result in a model that is less likely to overfit but may not fit the data as well. Similarly, fewer splines would result in a simpler model that is less likely to overfit but may not fit the data as well, while more splines would result in a more complex model that fits the data well but is likely to overfit. 

In our experiments, we have search domains that consist of both discrete and categorical hyperparameters. To perform optimization on these domains, we first cast each hyperparameter to its corresponding index in an array. Then, we convert these indices to floating-point numbers so that we can use continuous optimization algorithms. Finally, we round the output of the optimization algorithm to the nearest index to get the final hyperparameter configuration. This allows us to use continuous optimization algorithms on search domains that consist of discrete and categorical hyperparameters.

\begin{table}[ht]
\small
\renewcommand{\arraystretch}{1.3}
\centering
\caption{OpenML Task IDs used for HPOBench experiments. The table displays the total number of instances (combining the training and testing sets) (\#obs) and the number of features prior to any preprocessing (\#feat) for each dataset.}
\label{table:openML-task-ids}
\begin{tabular}{@{}llrrr@{}}
\toprule
Name & TID & \#obs & \#feat \\ \midrule
blood-transf.. & 10101 & 748 & 4 \\
vehicle & 53 & 846 & 18 \\
Australian & 146818 & 690 & 14 \\
car & 146821 & 1728 & 6 \\
phoneme & 9952 & 5404 & 5 \\
segment & 146822 & 2310 & 19 \\
credit-g & 31 & 1000 & 20 \\
kc1 & 3917 & 2109 & 22 \\
sylvine & 168912 & 5124 & 20 \\
kr-vs-kp & 3 & 3196 & 36 \\
jungle\_che.. & 167119 & 44819 & 6 \\
mfeat-factors & 12 & 2000 & 216 \\
shuttle & 146212 & 58000 & 9 \\
jasmine & 168911 & 2984 & 145 \\
cnae-9 & 9981 & 1080 & 856 \\
numerai28.6 & 167120 & 96320 & 21 \\
bank-mark.. & 14965 & 45211 & 16 \\
higgs & 146606 & 98050 & 28 \\
adult & 7592 & 48842 & 14 \\
nomao & 9977 & 34465 & 118 \\ \bottomrule
\end{tabular}
\end{table}

\begin{figure}
\begin{center}
  \includegraphics[width=\textwidth]{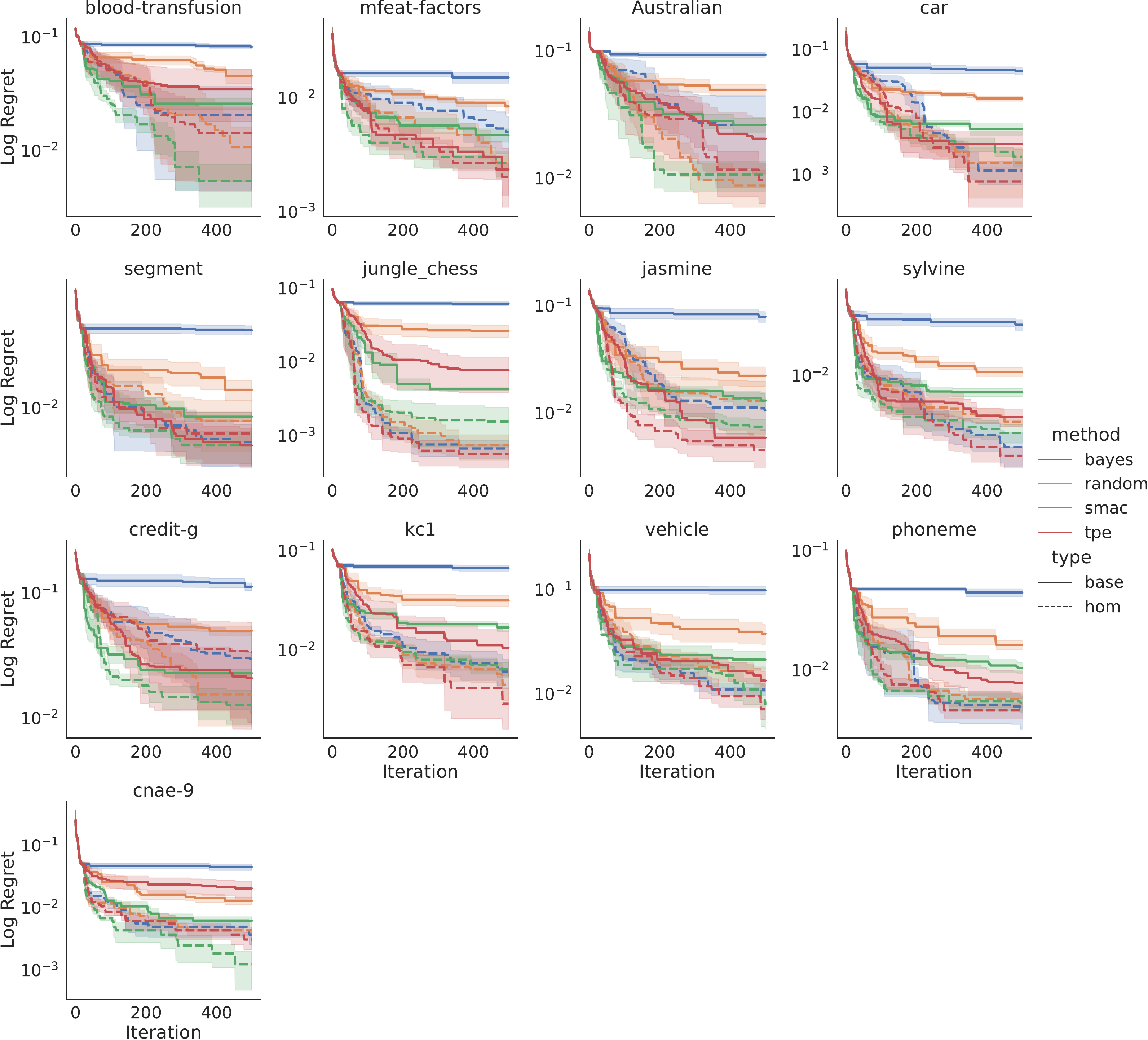}
  \caption{Comparison of all methods on 13 different tasks for the Random Forest Benchmark from the HPOBench suite. The mean and standard error of the regret at each iteration are displayed across 5 repetitions.}
  \label{plt:rf}
\end{center}
\end{figure}

\begin{table}[ht]
\small
\renewcommand{\arraystretch}{1.3}
\centering
\caption{Configuration spaces for the benchmarks included in HPOBench, with hyperparameters and their ranges for each model.}
\label{tab:hpobench}
\begin{tabular}{@{}llll@{}}
\toprule
Benchmark & Name & Range \\
\midrule
SVM & C & [2$^{-10}$, 2$^{10}$] \\
& gamma & [2$^{-10}$, 2$^{10}$] \\
\midrule
LogReg & alpha & [1e-05, 1.0] \\
& eta0 & [1e-05, 1.0] \\
\midrule
XGBoost & colsample\_bytree & [0.1, 1.0] \\
& eta & [2$^{-10}$, 1.0] \\
& max\_depth & [1, 50] \\
& reg\_lambda & [2$^{-10}$, 2$^{10}$] \\
\midrule
RandomForest & max\_depth & [1, 50] \\
& max\_features & [0.0, 1.0] \\
& min\_samples\_leaf & [1, 2] \\
& min\_samples\_split & [2, 128] \\
\midrule
MLP & alpha & [1.0e$^{-08}$, 1.0] \\
& batch\_size & [4, 256] \\
& depth & [1, 3] \\
& learning\_rate\_init & [1.0e$^{-05}$, 1.0] \\
& width & [16, 1024] \\
\bottomrule
\end{tabular}
\end{table}

\subsection{Results}

\subsubsection{HPOBench Classification Experiments}

\begin{figure}
\begin{center}
    \includegraphics[width=\textwidth]{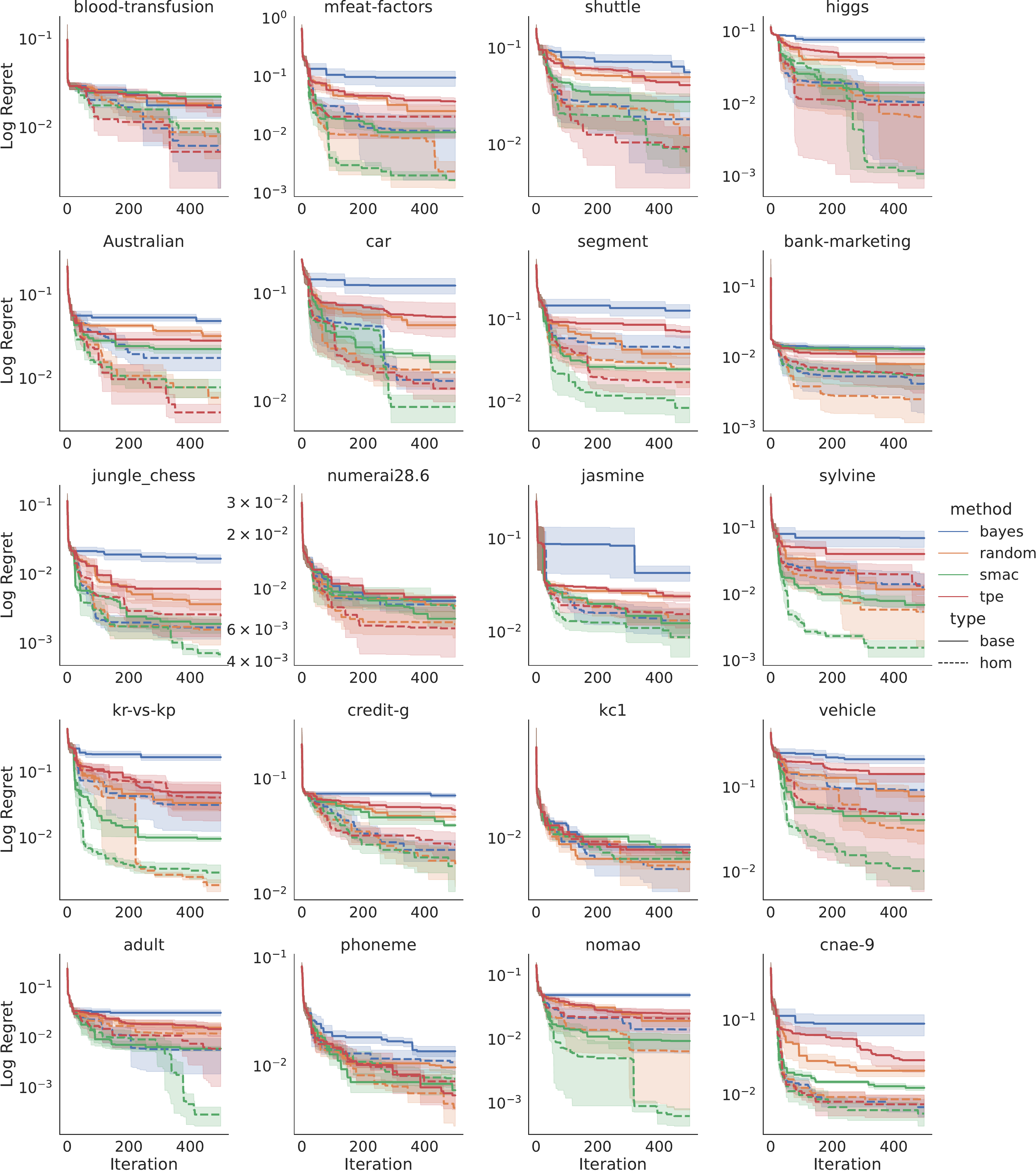}
    \caption{Comparison of all methods on 20 different tasks for the Logistic Regression Benchmark from the HPOBench suite. The mean and standard error of the regret at each iteration are displayed across 5 repetitions.}
    \label{plt:lr}
\end{center}
\end{figure}

\begin{figure}
\begin{center}
    \includegraphics[width=\textwidth]{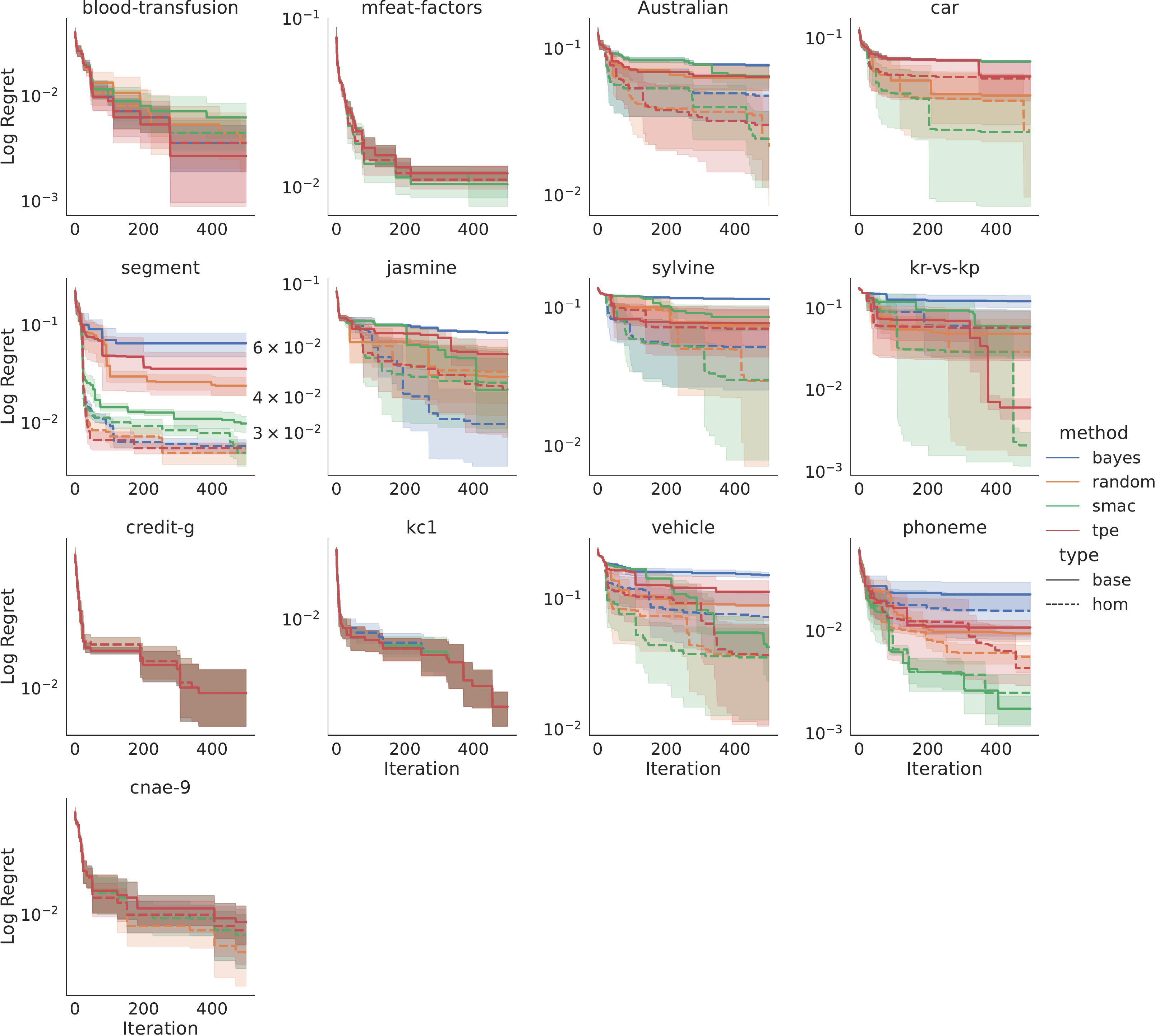}
    \caption{Comparison of all methods on 13 different tasks for the SVM Benchmark from the HPOBench suite. The mean and standard error of the regret at each iteration are displayed across 5 repetitions.}
    \label{plt:svm}
\end{center}
\end{figure}

\begin{figure}
\begin{center}
    \includegraphics[width=\textwidth]{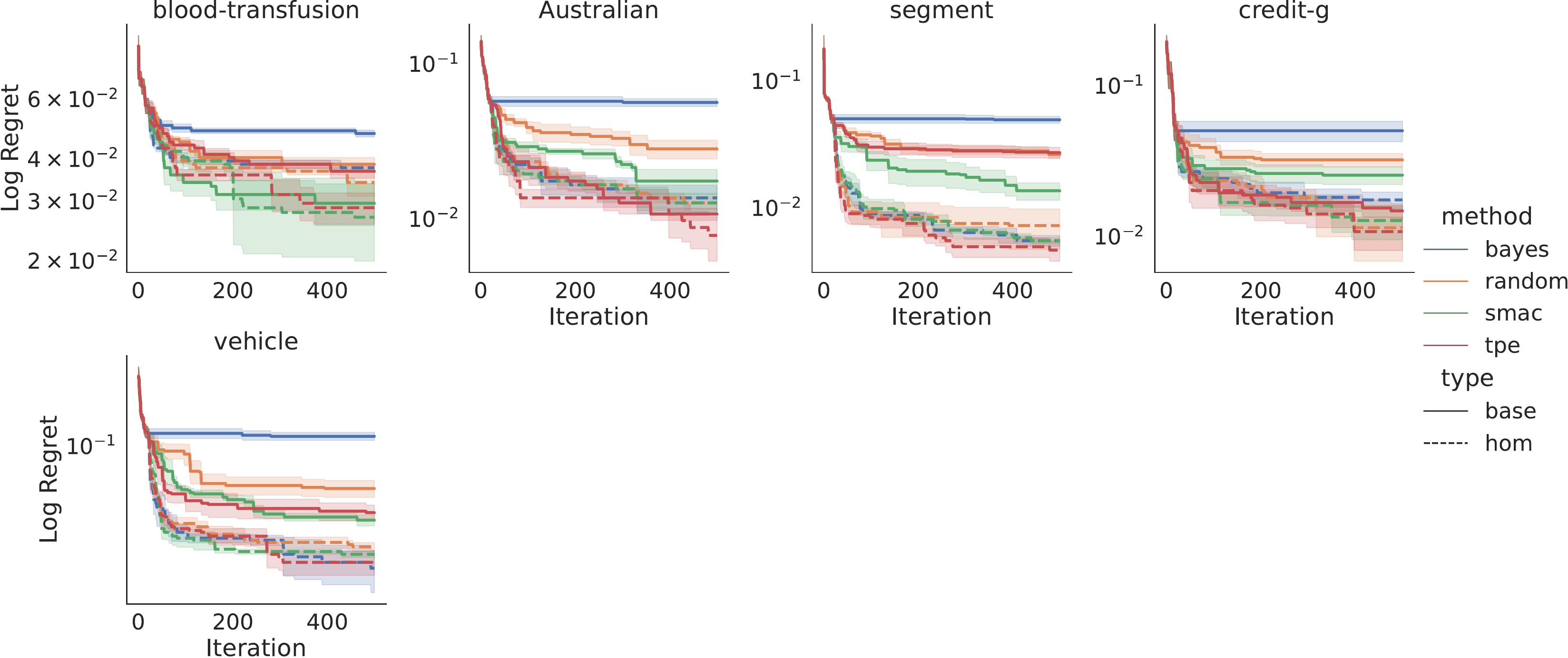}
    \caption{Comparison of all methods on 5 different tasks for the MLP Benchmark from the HPOBench suite. The mean and standard error of the regret at each iteration are displayed across 5 repetitions.}
    \label{plt:nn}
\end{center}
\end{figure}

\begin{figure}
\begin{center}
    \includegraphics[width=\textwidth]{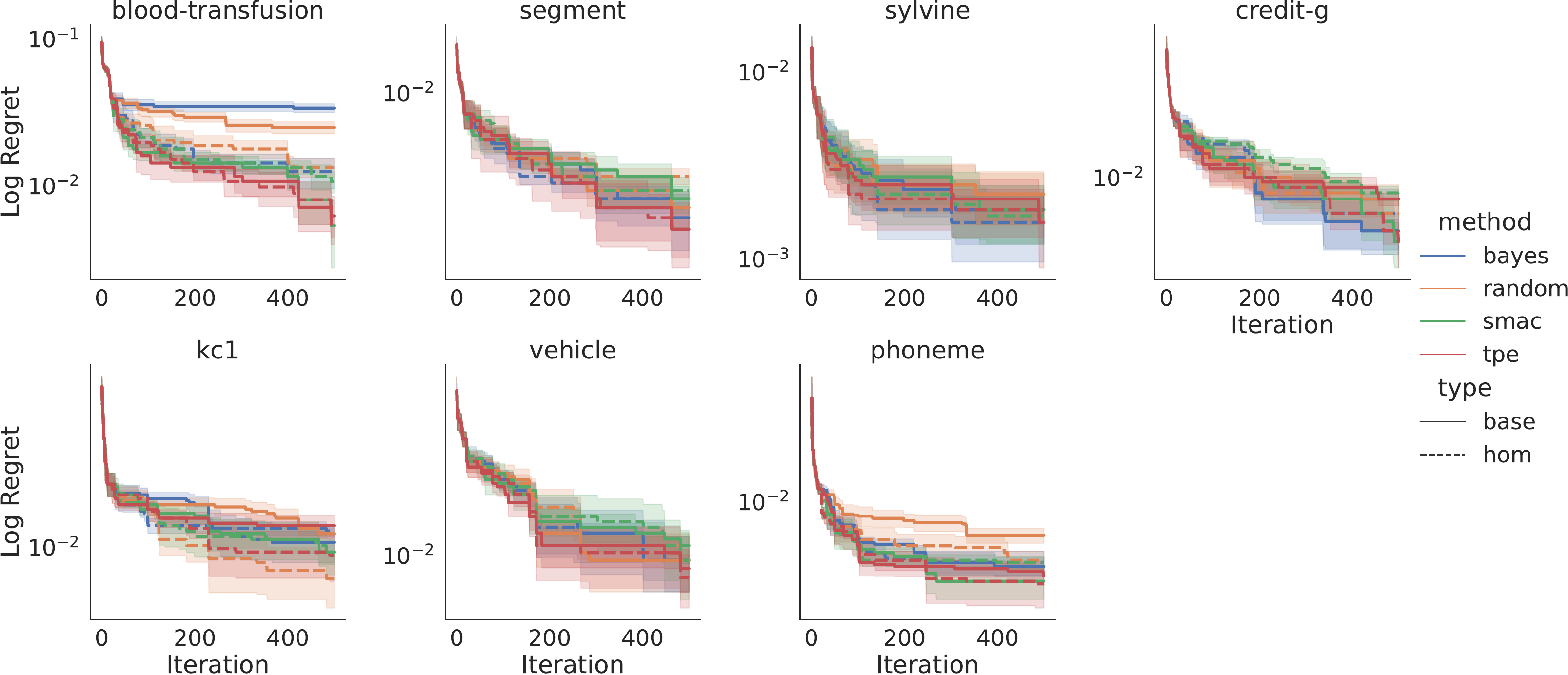}
    \caption{Comparison of all methods on 14 different tasks for the XGBoost Benchmark from the HPOBench suite. The mean and standard error of the regret at each iteration are displayed across 5 repetitions.}
    \label{plt:xgboost}
\end{center}
\end{figure}

We compare the performance of \HomOpt{} on the tuning of different sets of parameters on the SVM, Random Forest, Logistic Regression, MLP and XGBoost benchmarks from the HPOBench suite across multiple dataset tasks from the OpenML library. In all these experiments we use the same parameters for \HomOpt{} across all benchmarks and datasets. Figures \ref{plt:rf} - \ref{plt:xgboost} reflect the regret at each iteration for all the ML benchmarks. The regret plots illustrate the difference between the performance at each iteration compared to the best value found for the entire dataset. Thus, the lower the regret, the better the optimization algorithm is performing. 
A summary of the performance on the best observed validation accuracy is additionally included in Appendix \ref{sec:extended-val} where we see an improvement in minimizing the validation loss for a majority of the datasets in 4 of the 5 benchmarks.

In the Random Forest benchmark shown in Figure \ref{plt:rf}, for each of the 13 datasets, we can see a consistent trend where \HomOpt{} successfully improves the performance of all the base methods alone where the regret plots demonstrated a faster convergence to a better optima. The trials where \HomOpt{} augmented the base methodologies are indicated with the dashed line which have steeper slope compared to the base methods alone (indicated by the solid lines). Each of the datasets had varying number of instances and features as indicated in Table \ref{table:openML-task-ids}. The most significant improvement of \HomOpt{} over the base methods can actually be seen in the dataset with the highest number of features but relatively low number of instances (cnae-9). Random Forest has four hyperparameters (Table \ref{tab:hpobench}), and although the ranges may not be too wide, the combination and interaction of this space can make it challenging. The min samples leaf and min samples split hyperparameters have relatively narrow ranges, but the interaction between these hyperparameters and max depth can make the search space challenging. For example, setting min samples leaf or min samples split too small can lead to overfitting, especially when max depth is also large. On the other hand, setting these hyperparameters too large can lead to underfitting, especially when max depth is small.

Compared to the Random Forest benchmark, the SVM benchmark illustrated in Figure \ref{plt:svm} consists of only two hyperparameters, leading to a lower dimensionality in its search space. However, the search space for SVM is more extensive than that of the Random Forest benchmark, as both hyperparameters (C and gamma) have a range spanning from 2$^{-10}$ to 2$^{10}$. These ranges are significantly large, covering several orders of magnitude, which poses a challenge when navigating through the search space. \HomOpt{} demonstrates a relative boost over the base methodologies regarding in specific datasets such as the Australian, car, segment, jasmine, sylvine, and vehicle datasets. On the other hand, in datasets like blood-transfusion, mfeat-factors, credit-g, kc1, and cnae-9, the results across all methods, including those augmented with \HomOpt{}, are relatively similar. The search space in these datasets may contain complex interactions between hyperparameters, making it difficult for all methods to find the optimal configurations. Consequently, the convergence to the optimum in these cases may not be as significant when compared to the base methodologies for these datasets. While \HomOpt{} with the default parameters demonstrates an overall boost in many of the datasets concerning the overall optima, the convergence to the optimum is not as pronounced. In several cases, the results are comparable or only marginally better than the base methodologies. This suggests that \HomOpt{} might offer advantages in specific scenarios, but its performance may be influenced by a combination of factors, such as the inherent complexity of the datasets, the optimization algorithm's ability to navigate the search space, and the interactions between hyperparameters in the SVM benchmark.


The search space for the Logistic Regression benchmark is relatively straightforward, consisting of only two hyperparameters: alpha and eta0. These hyperparameters have a range of 1e-05 to 1.0, which is notably smaller compared to the broader ranges observed in SVM's hyperparameters. This simpler search space allows for more manageable exploration and optimization, as it does not involve as many complex interactions between hyperparameters as seen in other benchmarks. The impact of \HomOpt{} on the Logistic Regression benchmark (Figure \ref{plt:lr}) exhibits a trend similar to the results observed in the Random Forest benchmark. In both cases, the trials incorporating \HomOpt{} demonstrate an overall improvement in performance across all 20 datasets. Additionally, \HomOpt{} facilitates a faster convergence towards a lower optimal value, suggesting that the optimization algorithm is more efficient in navigating the search space and identifying better hyperparameter configurations than the base methodologies alone. This improved performance of \HomOpt{} in the Logistic Regression benchmark can be attributed to several factors. The relatively simple search space, with fewer hyperparameters and a smaller range, enables a more effective exploration of possible configurations. Additionally, the inherently lower complexity of the Logistic Regression model, compared to models with larger search spaces or more hyperparameters, could also contribute to the improved performance observed when using \HomOpt{}.

The MLP model has a larger search space with its five hyperparameters, leading to a greater number of possible configurations. The interactions between these hyperparameters are quite complex, further contributing to the challenge of optimizing this model. For instance, the depth and width of the network directly impact its capacity and complexity, while the alpha (L2 regularization) and learning rate init parameters control the model's generalization performance. Additionally, the batch size parameter influences not only the convergence speed but also the quality of the final solution. Navigating these intricate interactions within the MLP search space can be extremely challenging for HPO algorithms. Despite these challenges, \HomOpt{} has demonstrated a significant boost in performance compared to the base methodologies in 5 different datasets (Figure \ref{plt:nn}). This improvement is evidenced by lower optimal scores and faster convergence rates. Furthermore, the standard error of the regret in the MLP benchmark is consistently lower than that of the SVM benchmark across the five repetitions. This suggests that the optimization process for MLP is more stable and less prone to random fluctuations compared to SVM. This could be attributed to the fact that the MLP search space is more constrained compared to SVM due to the smaller range of hyperparameters, resulting in a more focused search. This, in turn, could lead to a more stable optimization process with a more prominent boost in the performance.

XGBoost has four hyperparameters: colsample bytree, eta, max depth, and reg lambda. The search space complexity is higher than that of SVM and Logistic Regression due to the increased number of hyperparameters. The ranges of these hyperparameters are quite different, with some having smaller ranges (e.g., colsample bytree) and others having larger ranges (e.g., max depth and reg lambda). Among all five of the benchmarks from HPOBench, \HomOpt{} demonstrated the least noticeable improvement in the XGBoost experiments. As seen by the regret plots in Figure \ref{plt:xgboost} the results among all the methods appeared to have a similar effect in regards to the convergence and the overall optimum. We can see a small boost in performace for random and bayes with \HomOpt{} in the blood-transfusion dataset which has only 4 features and 748 instances but otherwise a noticeable boost can not be seen. One possible reason why \HomOpt{} did not perform as well on the XGBoost benchmark compared to the other benchmarks could be due to the interactions between the four hyperparameters. The interactions may be more intricate and complex compared to the other benchmarks, making it more challenging for \HomOpt{} to effectively explore the search space and find the optimal solution. These hyperparameters are also interdependent with each other. For instance, a higher eta value may require a higher reg lambda value to counteract the increased learning rate. The ranges of the hyperparameters in XGBoost vary widely. For example, the colsample bytree parameter has a range of [0.1, 1.0], while the max depth and reg lambda parameters have ranges of [1, 50] and [2$^{-10}$, 2$^{10}$], respectively. This heterogeneity in the ranges could also contribute to the difficulty of exploring the search space efficiently.

\begin{table}
\centering
\small
\renewcommand{\arraystretch}{1.3}
\begin{tabular}{llll}
\toprule
\textbf{Variable} & \textbf{Description} & \textbf{Domain} \\ \midrule
$\mathcal{D}$ & Local perturbation factor & [5, 5e-1, 5e-2, 5e-3, 5e-4, 5e-5] \\ 
$\mathcal{W}$ & Number of warm-up samples & [10, 30, 50, 70, 90] \\ 
$\mathcal{N}$ & Number of minimization steps to compute homotopy & [3, 6, 9] \\ 
$k$ & Fraction of completed trials to train the GAMs & [0.2, 0.4, 0.6, 0.8, 1] \\ 
\bottomrule
\end{tabular}
\caption{\HomOpt{} parameters and domain spaces for ablation study on XGBoost benchmark.}
\label{tab:hyperparams}
\end{table}

\begin{figure}[!htb]
    \centering
    \includegraphics[width=1.0\textwidth]{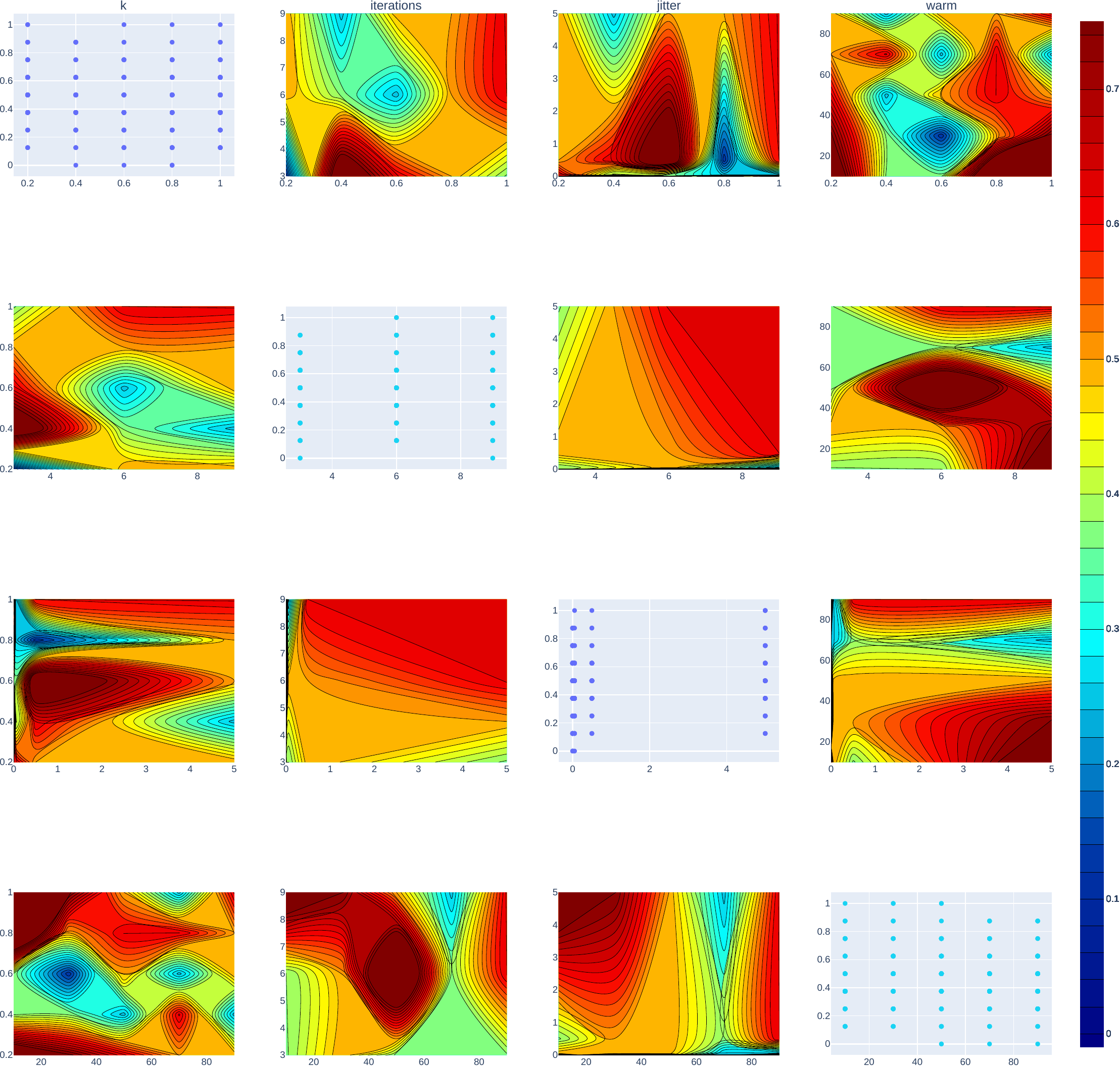}
    \caption{Contour plots visualizing the interactions between pairs of method parameters with SMAC samples on XGBoost sensitivity study.  In this plot $k$ represents the proportion of trial data used to train one of the GAMs, $iterations$ are the number of minimizations to compute the homotopy, $jitter$ refers to the distance threshold as the local perturbation search around the observed minimum and $warm$ represents the number of samples used for the surrogate approximation. Changes in the method parameter values can affect the optimization performance, where certain regions have higher performance (indicated by a lower loss value which are the blue regions)}
    \label{fig:smac_contour_xgboost}
\end{figure}

Since \HomOpt{} demonstrated minimal improvement on the XGBoost benchmark, we performed an additional study on one of the datasets to illustrate the sensitivity of parameters within \HomOpt{} on the performance of the optimization. On dataset credit-g, we run \HomOpt{} for each of the base sampling methods (Bayes, Random, SMAC, TPE), with the method parameters and domains indicated in Table \ref{tab:hyperparams}. This search was done over 100 iterations for a single seed for each combination of method parameters. The studies visualized in the contour plot (Figure \ref{fig:smac_contour_xgboost}) with SMAC samples, revealed that the performance of the optimization using \HomOpt{} is sensitive to the choice of corresponding method parameters and domains. In our experiments our default method parameters use a $k$ of 0.5, five iterations to compute the homotopy, a jitter strength of 0.005 and 20 warm up samples. The contour plot demonstrates how changes in these method parameter values can affect the optimization performance, where certain regions have higher performance (indicated by a lower loss value which are the blue regions). In this case, a higher $k$ value of 0.6 and more warm samples (30+) resulted in higher performance for this dataset.

\subsubsection{Open-Set Benchmarks}





\begin{figure}
\begin{center}
  \includegraphics[width=\textwidth]{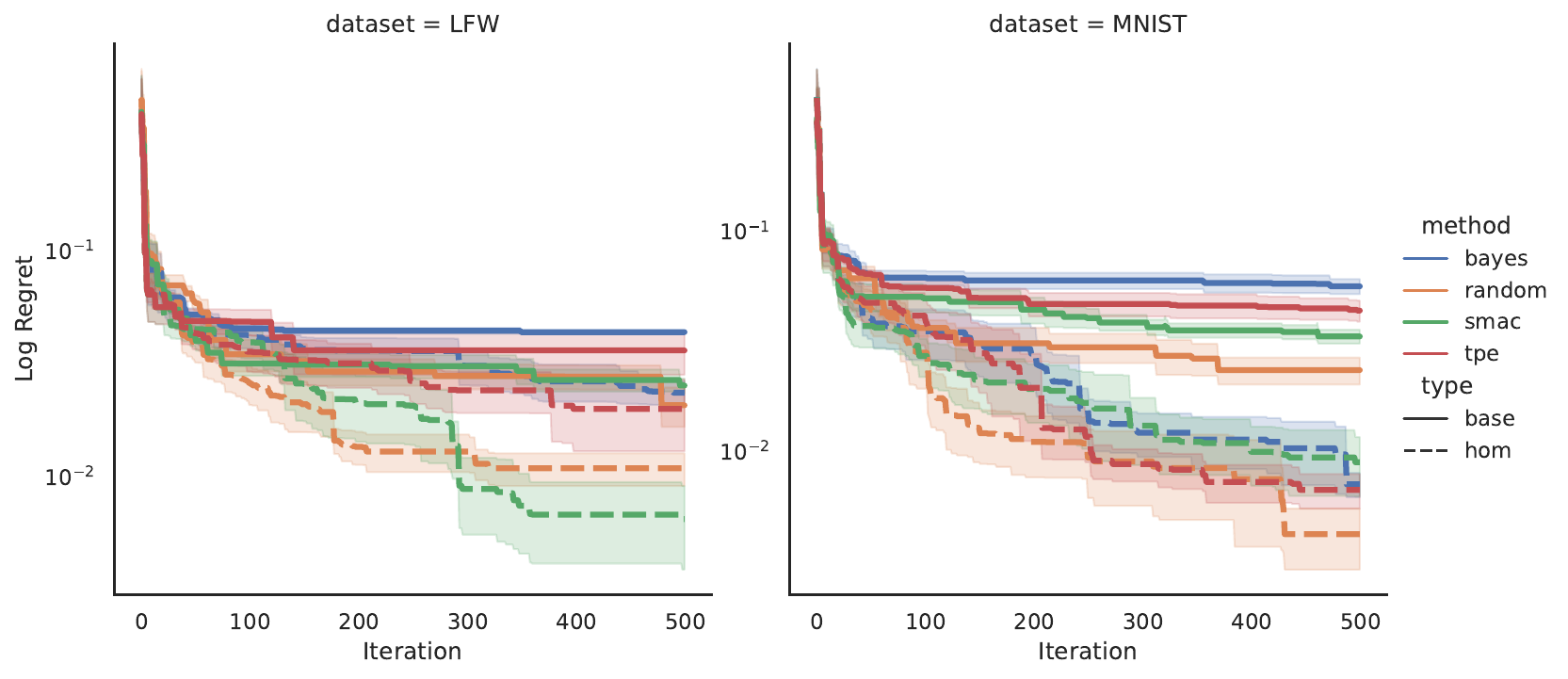}
  \caption{Comparison of all methods on MNIST and LFW for the open-set benchmarks. The mean and standard error of the regret at each iteration are displayed across 5 repetitions. \HomOpt{} boosts the performance of all the methods for both of the datasets.}
  \label{plt:evm-regret}
\end{center}
\end{figure}

Optimizing the EVM model can be challenging due to the complexity of its hyperparameters and the high-dimensional nature of the data it handles. The EVM model has five hyperparameters, each with its own domain and range, which must be tuned to achieve optimal performance. The tailsize parameter, in particular, can be difficult to optimize, as it determines the number of negative samples used to estimate the model parameters. This parameter can have a significant impact on the model's performance, but it is also highly dependent on the characteristics of the input data. Additionally, the distance function parameter can be difficult to optimize because it affects the way the model computes distances between samples, and this can have a significant impact on the model's accuracy. Finding the right combination of hyperparameters to optimize the EVM model can be a time-consuming and challenging task for any HPO algorithm. In Figure \ref{plt:evm-regret}, we evaluate the performance of \HomOpt{} in tuning the 5 hyperparameters of an open-set recognition model, the EVM on two separate datasets over 5 random seeds. \HomOpt{} again demonstrates a faster convergence to a better optimum value for both datasets, boosting all of the methods that were tested and also on average found a better objective value (observed $F1$ score). Compared to some of the other models in HPOBench, the EVM has a relatively small and constrained hyperparameter search space. The threshold, tailsize, cover threshold, and distance multiplier parameters all have ranges between 0 and 1, while the distance function is limited to two options: Cosine or Euclidean distance. This limited search space may partially explain why \HomOpt{} was able to perform well on the EVM benchmark.

%% file: 06conclusion.tex
\section{Conclusion \& Limitations}


We introduced \HomOpt{}, a hyperparameter optimization search strategy that builds a continuous deformation between GAM surrogate models and employs homotopy methods to track local minima along the deformation. \HomOpt{} effectively identifies approximate regions of interest on the hyperparameter surface, allowing for efficient space exploration and faster convergence to an optimal solution. We demonstrated \HomOpt{}'s compatibility with various popular HPO methods, as it consistently speeds up their convergence and often finds better optima across numerous benchmarks. These benchmarks represent a wide range of challenging optimization problems, showcasing the rigor of our evaluation. Furthermore, \HomOpt{} is adaptable to any HPO method and exhibits improved performance with faster convergence for both closed- and open-set models, without requiring strong assumptions about the objective. Future work will explore extending \HomOpt{} to other optimization scenarios, incorporating additional surrogate models, and investigating its potential in multi-objective optimization problems. Ultimately, \HomOpt{} holds promise in advancing HPO methods and facilitating the development of more robust and efficient models across various applications.

%% file: 07limitations.tex

Although \HomOpt{} demonstrated robust performance across multiple benchmarks, there are a few limitations to consider. For this work, we utilized a GAM as a surrogate model which in itself contains hyperparameters. In all the experiments, recall our GAM model surrogates use a penalty term on the smooth functions of $10^{-4}$ and 25 splines. \HomOpt{} also has configurable parameters that are difficult to intuitively choose posing an additional challenge in driving the optimization. Potential ways to mitigate this would be to adaptively select all of the meta-parameters with methods such as reinforcement learning \citep{sutton2018reinforcement}. The choice of surrogate model may also affect the performance of the method. Although this remains outside of the scope of this paper, alternatives to the GAM include random forests \citep{breiman2001random}, Bayesian networks \citep{ghahramani2006learning}, gradient boosting machines \citep{friedman2001greedy}, among many others \citep{bhosekar2018advances}. However it is important to note that the GAM provides a more interpretable surrogate model against these suggested alternative approaches. 

In the current configuration of \HomOpt{}, only one optimal point is found. However, this can be modified by utilizing a different surrogate which tracks the change of multiple minima at spread out regions. This can also be modified by taking the number of points within the top $\alpha\%$ instead of utilizing a single point. We also consider the adaption of \HomOpt{} for other applications key to hyperparameter optimization such as incorporating domain knowledge. \HomOpt{} provide a framework integrating the the use of homotopies to track the transition between minimums which can enable the incorporation of domain knowledge in the form of constraints or heuristics that better guide the optimization process.

%% file: 08acknowledgements.tex
\section*{Acknowledgements}

 This work was funded by DEVCOM Army Research Laboratory under cooperative agreement, W911NF-20-2-0218.

%% file: appendix.tex
\section{Extended Results}
\label{sec:extended-results}
\subsection{EVM Results}
\label{sec:extended-evm}
Here we include additional results from the experiments. In the EVM experiments, we report the corresponding Normalized Mutual Information (NMI) score for each experiment which measures the similarity between the predicted label and the true class label as well.

\begin{equation}
\text{NMI}(C, T) = \frac{2 \times I(C, T)}{H(C) + H(T)}
\label{eqn:nmi}
\end{equation}

\begin{equation}
\text{H}(C) = - \sum_{c} p(c) \log_2 p(c)
\end{equation}

\begin{equation}
\text{H}(T) = - \sum_{t} p(t) \log_2 p(t)
\end{equation}

\input{tables/EVM_extra}

\subsection{HPOBench}
\label{sec:extended-hpobench}
In this section, we present the additional results for the HPOBench experiments, specifically focusing on the validation and test scores. The validation and test scores showcase the average percent improvement and standard error achieved over 5 trials for each dataset experiment within the benchmark. 
\subsubsection{Validations Scores}
\label{sec:extended-val}
\input{tables/SVMBenchmark_table}
\input{tables/RandomForestBenchmark_table}
\input{tables/LRBenchmark_table}
\input{tables/NNBenchmark_table}
\input{tables/XGBoostBenchmark_table}

\clearpage
\subsubsection{Test Scores}
\input{tables/RandomForestBenchmark_test_loss_table}
\input{tables/SVMBenchmark_test_loss_table}
\input{tables/LRBenchmark_test_loss_table}
\input{tables/NNBenchmark_test_loss_table}
\input{tables/XGBoostBenchmark_test_loss_table}

%% file: tables/EVM_extra.tex
Table~\ref{tab:table-exp-EVM-MNIST} reported the validation and testing score for MNIST data set experiments for open set recognition. In general, we can see the results for \HomOpt{} are improved against all the base strategies random, TPE, Bayes, and SMAC. We can see consistent results (see Table~\ref{tab:table-exp-EVM-LFW}) in the LFW experiment improves among all the base methods. It is noticeable that the highest validation and testing scores for MNIST data set achieves from the \HomOpt{} with random seeds used for warm-up, while LFW data set achieves the highest scores from the \HomOpt{} with SMAC seeds used for warm-up trials. 
\begin{table}[ht!]
\resizebox{\textwidth}{!}{%
\begin{tabular}{@{}lcccccc@{}}
\toprule
\multicolumn{1}{c}{\textbf{Method}} &
  \textbf{Validation F1} &
  \textbf{Validation Accuracy} &
  \textbf{Validation NMI} &
  \textbf{Testing F1} &
  \textbf{Testing Accuracy} &
  \textbf{Testing NMI} \\ \midrule
\multicolumn{1}{c}{\textbf{Random}} &
  \multicolumn{1}{c}{0.8982 $\pm$ 0.006} &
  \multicolumn{1}{c}{0.8958 $\pm$ 0.007} &
  \multicolumn{1}{c}{0.7547 $\pm$ 0.014} &
  \multicolumn{1}{c}{0.8988 $\pm$ 0.005} &
  \multicolumn{1}{c}{0.8969 $\pm$ 0.007} &
  \multicolumn{1}{c}{0.7579 $\pm$ 0.013} \\
\multicolumn{1}{c}{\textbf{PSHHO+Random}} &
  \multicolumn{1}{c}{\textbf{0.9160 $\pm$ 0.003}} &
  \multicolumn{1}{c}{\textbf{0.9127 $\pm$ 0.004}} &
  \multicolumn{1}{c}{\textbf{0.7889 $\pm$ 0.008}} &
  \multicolumn{1}{c}{\textbf{0.9039 $\pm$ 0.006}} &
  \multicolumn{1}{c}{\textbf{0.9022 $\pm$ 0.007}} &
  \multicolumn{1}{c}{\textbf{0.7657 $\pm$ 0.012}} \\
\multicolumn{1}{c}{\textbf{TPE}} &
  \multicolumn{1}{c}{0.8889 $\pm$ 0.011} &
  \multicolumn{1}{c}{0.8859 $\pm$ 0.011} &
  \multicolumn{1}{c}{0.7431 $\pm$ 0.022} &
  \multicolumn{1}{c}{0.8788 $\pm$ 0.004} &
  \multicolumn{1}{c}{0.8766 $\pm$ 0.004} &
  \multicolumn{1}{c}{0.7287 $\pm$ 0.009} \\
\multicolumn{1}{c}{\textbf{PSHHO+TPE}} &
  \multicolumn{1}{c}{0.9140 $\pm$ 0.003} &
  \multicolumn{1}{c}{0.9106 $\pm$ 0.004} &
  \multicolumn{1}{c}{0.7838 $\pm$ 0.005} &
  \multicolumn{1}{c}{0.8982 $\pm$ 0.007} &
  \multicolumn{1}{c}{0.8964 $\pm$ 0.006} &
  \multicolumn{1}{c}{0.7579 $\pm$ 0.012} \\
\multicolumn{1}{c}{\textbf{Bayes}} &
  \multicolumn{1}{c}{0.8771 $\pm$ 0.008} &
  \multicolumn{1}{c}{0.8749 $\pm$ 0.006} &
  \multicolumn{1}{c}{0.7200 $\pm$ 0.015} &
  \multicolumn{1}{c}{0.8669 $\pm$ 0.010} &
  \multicolumn{1}{c}{0.8655 $\pm$ 0.008} &
  \multicolumn{1}{c}{0.7054 $\pm$ 0.018} \\
\multicolumn{1}{c}{\textbf{PSHHO+Bayes}} &
  \multicolumn{1}{c}{0.9132 $\pm$ 0.002} &
  \multicolumn{1}{c}{0.9092 $\pm$ 0.003} &
  \multicolumn{1}{c}{0.7793 $\pm$ 0.003} &
  \multicolumn{1}{c}{0.8995 $\pm$ 0.003} &
  \multicolumn{1}{c}{0.8981 $\pm$ 0.004} &
  \multicolumn{1}{c}{0.7583 $\pm$ 0.004} \\
\multicolumn{1}{c}{\textbf{SMAC}} &
  \multicolumn{1}{c}{0.9034 $\pm$ 0.005} &
  \multicolumn{1}{c}{0.9001 $\pm$ 0.006} &
  \multicolumn{1}{c}{0.7650 $\pm$ 0.013} &
  \multicolumn{1}{c}{0.8924 $\pm$ 0.006} &
  \multicolumn{1}{c}{0.8907 $\pm$ 0.007} &
  \multicolumn{1}{c}{0.7468 $\pm$ 0.013} \\
\multicolumn{1}{c}{\textbf{PSHHO+SMAC}} &
  \multicolumn{1}{c}{0.9110 $\pm$ 0.006} &
  \multicolumn{1}{c}{0.9081 $\pm$ 0.005} &
  \multicolumn{1}{c}{0.7797 $\pm$ 0.009} &
  \multicolumn{1}{c}{0.8992 $\pm$ 0.004} &
  \multicolumn{1}{c}{0.8976 $\pm$ 0.005} &
  \multicolumn{1}{c}{0.7599 $\pm$ 0.008} \\
 \bottomrule
\end{tabular}%
}
\caption{Average validation/testing score comparison for each base method and corresponding homotopy approach across 5 separate seeds for the EVM experiments with MNIST dataset.}
\label{tab:table-exp-EVM-MNIST}
\end{table}

\begin{table}[ht!]
\resizebox{\textwidth}{!}{%
\begin{tabular}{@{}lcccccc@{}}
\toprule
\multicolumn{1}{c}{\textbf{Method}} &
  \textbf{Validation F1} &
  \textbf{Validation Accuracy} &
  \textbf{Validation NMI} &
  \textbf{Testing F1} &
  \textbf{Testing Accuracy} &
  \textbf{Testing NMI} \\ \midrule
\multicolumn{1}{c}{\textbf{Random}} &
  \multicolumn{1}{c}{0.8556 $\pm$ 0.009} &
  \multicolumn{1}{c}{0.9659 $\pm$ 0.004} &
  \multicolumn{1}{c}{0.8254 $\pm$ 0.019} &
  \multicolumn{1}{c}{0.8191 $\pm$ 0.058} &
  \multicolumn{1}{c}{0.9596 $\pm$ 0.028} &
  \multicolumn{1}{c}{0.7927 $\pm$ 0.011} \\
\multicolumn{1}{c}{\textbf{PSHHO+Random}} &
  \multicolumn{1}{c}{0.8654 $\pm$ 0.004} &
  \multicolumn{1}{c}{0.9673 $\pm$ 0.001} &
  \multicolumn{1}{c}{0.8309 $\pm$ 0.009} &
  \multicolumn{1}{c}{\textbf{0.8248 $\pm$ 0.015}} &
  \multicolumn{1}{c}{0.9610 $\pm$ 0.002} &
  \multicolumn{1}{c}{0.7981 $\pm$ 0.011} \\
\multicolumn{1}{c}{\textbf{TPE}} &
  \multicolumn{1}{c}{0.8400 $\pm$ 0.018} &
  \multicolumn{1}{c}{0.9618 $\pm$ 0.006} &
  \multicolumn{1}{c}{0.8060 $\pm$ 0.026} &
  \multicolumn{1}{c}{0.7992 $\pm$ 0.026} &
  \multicolumn{1}{c}{0.9554 $\pm$ 0.005} &
  \multicolumn{1}{c}{0.7729 $\pm$ 0.024} \\
\multicolumn{1}{c}{\textbf{PSHHO+TPE}} &
  \multicolumn{1}{c}{0.8563 $\pm$ 0.016} &
  \multicolumn{1}{c}{0.9632 $\pm$ 0.007} &
  \multicolumn{1}{c}{0.8140 $\pm$ 0.030} &
  \multicolumn{1}{c}{0.8175 $\pm$ 0.014} &
  \multicolumn{1}{c}{0.9571 $\pm$ 0.005} &
  \multicolumn{1}{c}{0.7844 $\pm$ 0.023} \\
\multicolumn{1}{c}{\textbf{Bayes}} &
  \multicolumn{1}{c}{0.8426 $\pm$ 0.007} &
  \multicolumn{1}{c}{0.9617 $\pm$ 0.002} &
  \multicolumn{1}{c}{0.8049 $\pm$ 0.010} &
  \multicolumn{1}{c}{0.7919 $\pm$ 0.013} &
  \multicolumn{1}{c}{0.9542 $\pm$ 0.003} &
  \multicolumn{1}{c}{0.7657 $\pm$ 0.014} \\
\multicolumn{1}{c}{\textbf{PSHHO+Bayes}} &
  \multicolumn{1}{c}{0.8622 $\pm$ 0.008} &
  \multicolumn{1}{c}{0.9671 $\pm$ 0.003} &
  \multicolumn{1}{c}{0.8288 $\pm$ 0.016} &
  \multicolumn{1}{c}{0.8161 $\pm$ 0.015} &
  \multicolumn{1}{c}{0.9589 $\pm$ 0.003} &
  \multicolumn{1}{c}{0.78855 $\pm$ 0.015} \\
\multicolumn{1}{c}{\textbf{SMAC}} &
  \multicolumn{1}{c}{0.8510 $\pm$ 0.004} &
  \multicolumn{1}{c}{0.9657 $\pm$ 0.003} &
  \multicolumn{1}{c}{0.8241 $\pm$ 0.014} &
  \multicolumn{1}{c}{0.8218 $\pm$ 0.016} &
  \multicolumn{1}{c}{0.9602 $\pm$ 0.003} &
  \multicolumn{1}{c}{0.7951 $\pm$ 0.003} \\
\multicolumn{1}{c}{\textbf{PSHHO+SMAC}} &
  \multicolumn{1}{c}{\textbf{0.8698 $\pm$ 0.005}} &
  \multicolumn{1}{c}{\textbf{0.9699 $\pm$ 0.001}} &
  \multicolumn{1}{c}{\textbf{0.8433 $\pm$ 0.007}} &
  \multicolumn{1}{c}{0.8231 $\pm$ 0.010} &
  \multicolumn{1}{c}{\textbf{0.9613 $\pm$ 0.002}} &
  \multicolumn{1}{c}{\textbf{0.8001 $\pm$ 0.009}} \\
 \bottomrule
\end{tabular}%
}
\caption{Average validation/testing scores comparison for each base method and corresponding homotopy approach  across 5 separate seeds for the EVM experiment with LFW dataset.}
\label{tab:table-exp-EVM-LFW}
\end{table}

%% file: tables/SVMBenchmark_table.tex
\begin{table}
\centering
\caption{\textbf{Average percent improvement and standard error for the best observed loss over five trials of \HomOpt{} over the base methodology for each dataset and method in the SVM Benchmark. Bold values indicate where \HomOpt{} outperforms the base methodology.}}
\label{tab:f{SVM_BENCHMARK}}
\begin{tabular}{ccccc}
\toprule
\multirow{2}{*}{Dataset} & \multicolumn{4}{c}{Method}             \\
 & \textbf{bayes} &             \textbf{random} &              \textbf{smac} &                \textbf{tpe}                    \\
\midrule
10101   &    \textbf{5.22} $\pm$ 1.63 &    \textbf{4.35} $\pm$ 2.38 &    \textbf{8.33} $\pm$ 1.32 &    \textbf{5.22} $\pm$ 2.54 \\
12      &    \textbf{1.05} $\pm$ 0.43 &    \textbf{1.05} $\pm$ 0.43 &    \textbf{1.63} $\pm$ 0.94 &    \textbf{1.40} $\pm$ 0.47 \\
146818  &  \textbf{29.52} $\pm$ 17.26 &  \textbf{55.24} $\pm$ 12.74 &  \textbf{52.38} $\pm$ 12.78 &          -40.00 $\pm$ 44.97 \\
146821  &            -3.24 $\pm$ 1.01 &  \textbf{54.29} $\pm$ 26.30 &  \textbf{57.84} $\pm$ 25.82 &  \textbf{16.76} $\pm$ 20.82 \\
146822  &   \textbf{67.24} $\pm$ 1.09 &   \textbf{40.65} $\pm$ 2.62 &            -8.24 $\pm$ 5.46 &   \textbf{47.78} $\pm$ 1.36 \\
168911  &   \textbf{21.87} $\pm$ 5.96 &          -40.45 $\pm$ 11.20 &   \textbf{3.23} $\pm$ 10.08 &   \textbf{15.11} $\pm$ 8.84 \\
168912  &  \textbf{38.63} $\pm$ 16.36 &  \textbf{45.12} $\pm$ 15.89 &  \textbf{47.43} $\pm$ 14.96 &  \textbf{27.10} $\pm$ 16.45 \\
3       &  \textbf{56.34} $\pm$ 25.81 &   \textbf{9.68} $\pm$ 85.50 &   \textbf{98.47} $\pm$ 0.68 &        -587.50 $\pm$ 416.03 \\
31      &           -10.00 $\pm$ 5.39 &           -10.00 $\pm$ 5.39 &           -10.00 $\pm$ 5.39 &           -10.00 $\pm$ 5.39 \\
3917    &    \textbf{2.50} $\pm$ 1.67 &    \textbf{2.50} $\pm$ 1.67 &    \textbf{2.50} $\pm$ 1.67 &    \textbf{2.50} $\pm$ 1.67 \\
53      &  \textbf{41.54} $\pm$ 17.77 &  \textbf{28.67} $\pm$ 22.20 &   \textbf{0.00} $\pm$ 28.69 &          -33.75 $\pm$ 42.09 \\
9952    &          -14.86 $\pm$ 17.23 &            -1.59 $\pm$ 4.29 &            -1.72 $\pm$ 3.53 &            -1.64 $\pm$ 3.77 \\
9981    &            -2.75 $\pm$ 1.27 &            -2.00 $\pm$ 0.94 &            -2.75 $\pm$ 1.27 &            -3.00 $\pm$ 1.40 \\
\bottomrule
\end{tabular}
\label{table:svm}
\end{table}

%% file: tables/RandomForestBenchmark_table.tex
\begin{table}
\centering
\caption{\textbf{Average percent improvement and standard error for the best observed loss over five trials of \HomOpt{} over the base methodology for each dataset and method in the Random Forest benchmark. Bold values indicate where \HomOpt{} outperforms the base methodology.}}
\label{tab:rf-benchmark}
\begin{tabular}{ccccc}
\toprule
\multirow{2}{*}{Dataset} & \multicolumn{4}{c}{Method}             \\
 & \textbf{bayes} &             \textbf{random} &              \textbf{smac} &                \textbf{tpe}                    \\
\midrule
10101   &   \textbf{35.12} $\pm$ 8.78 &   \textbf{28.24} $\pm$ 3.55 &   \textbf{17.14} $\pm$ 1.75 &            -5.00 $\pm$ 9.26 \\
12      &  \textbf{60.00} $\pm$ 10.95 &   \textbf{35.00} $\pm$ 6.12 &   \textbf{48.00} $\pm$ 8.00 &         -120.00 $\pm$ 37.42 \\
146818  &  \textbf{61.82} $\pm$ 16.61 &   \textbf{63.08} $\pm$ 4.49 &   \textbf{42.22} $\pm$ 6.48 &  \textbf{16.67} $\pm$ 11.79 \\
146821  &   \textbf{93.04} $\pm$ 1.06 &   \textbf{80.00} $\pm$ 6.48 &   \textbf{0.00} $\pm$ 15.81 &          -40.00 $\pm$ 24.49 \\
146822  &   \textbf{69.70} $\pm$ 3.95 &  \textbf{38.95} $\pm$ 11.36 &   \textbf{24.62} $\pm$ 7.46 &   \textbf{3.64} $\pm$ 16.91 \\
167119  &   \textbf{46.36} $\pm$ 0.15 &   \textbf{34.72} $\pm$ 0.27 &    \textbf{3.99} $\pm$ 1.19 &    \textbf{0.59} $\pm$ 0.30 \\
168911  &   \textbf{54.29} $\pm$ 2.81 &   \textbf{22.32} $\pm$ 8.68 &   \textbf{13.21} $\pm$ 2.00 &            -0.00 $\pm$ 2.83 \\
168912  &   \textbf{73.75} $\pm$ 2.00 &   \textbf{40.00} $\pm$ 1.17 &   \textbf{28.00} $\pm$ 3.58 &   \textbf{14.74} $\pm$ 3.07 \\
31      &  \textbf{45.71} $\pm$ 13.24 &   \textbf{38.00} $\pm$ 6.80 &   \textbf{15.24} $\pm$ 5.51 &          -41.18 $\pm$ 42.78 \\
3917    &   \textbf{61.61} $\pm$ 2.00 &   \textbf{49.33} $\pm$ 1.91 &   \textbf{29.41} $\pm$ 2.08 &   \textbf{24.83} $\pm$ 2.76 \\
53      &   \textbf{64.44} $\pm$ 1.04 &   \textbf{42.50} $\pm$ 3.33 &            -9.09 $\pm$ 7.61 &    \textbf{1.67} $\pm$ 3.12 \\
9952    &   \textbf{63.68} $\pm$ 1.95 &   \textbf{29.18} $\pm$ 2.22 &   \textbf{25.26} $\pm$ 2.58 &    \textbf{1.43} $\pm$ 2.45 \\
9981    &   \textbf{81.67} $\pm$ 3.12 &   \textbf{40.00} $\pm$ 6.12 &  \textbf{30.00} $\pm$ 12.25 &   \textbf{60.00} $\pm$ 6.32 \\
\bottomrule
\end{tabular}
\label{table:rf}
\end{table}

%% file: tables/LRBenchmark_table.tex
\begin{table}
\centering
\caption{\textbf{Average percent improvement and standard error for the best observed loss over five trials of \HomOpt{} over the base methodology for each dataset and method in the Logistic Regression Benchmark. Bold values indicate where \HomOpt{} outperforms the base methodology.}}
\label{tab:lr-benchmark}
\begin{tabular}{ccccc}
\toprule
\multirow{2}{*}{Dataset} & \multicolumn{4}{c}{Method}             \\
 & \textbf{bayes} &             \textbf{random} &              \textbf{smac} &                \textbf{tpe}                    \\
\midrule
10101   &    \textbf{9.41} $\pm$ 1.44 &   \textbf{6.40} $\pm$ 1.60 &    \textbf{4.08} $\pm$ 1.12 &    \textbf{7.60} $\pm$ 1.47 \\
12      &  \textbf{78.79} $\pm$ 18.21 &  \textbf{90.00} $\pm$ 4.84 &   \textbf{75.00} $\pm$ 0.00 &  \textbf{35.79} $\pm$ 37.19 \\
146212  &  \textbf{39.34} $\pm$ 15.76 &  \textbf{46.21} $\pm$ 7.74 &   \textbf{37.63} $\pm$ 5.04 &   \textbf{44.88} $\pm$ 7.69 \\
146606  &   \textbf{17.02} $\pm$ 2.16 &   \textbf{4.57} $\pm$ 1.40 &    \textbf{6.01} $\pm$ 0.04 &    \textbf{5.94} $\pm$ 2.32 \\
146818  &   \textbf{16.88} $\pm$ 3.37 &  \textbf{19.33} $\pm$ 0.67 &   \textbf{12.14} $\pm$ 1.43 &   \textbf{20.67} $\pm$ 0.67 \\
146821  &   \textbf{29.49} $\pm$ 1.67 &  \textbf{27.44} $\pm$ 3.82 &   \textbf{16.77} $\pm$ 2.14 &            -3.46 $\pm$ 3.24 \\
146822  &   \textbf{7.93} $\pm$ 10.59 &   \textbf{1.91} $\pm$ 7.37 &    \textbf{9.21} $\pm$ 2.23 &   \textbf{25.00} $\pm$ 3.17 \\
14965   &    \textbf{4.33} $\pm$ 2.44 &   \textbf{4.82} $\pm$ 1.30 &    \textbf{2.91} $\pm$ 2.01 &    \textbf{3.03} $\pm$ 2.28 \\
167119  &    \textbf{2.96} $\pm$ 0.17 &   \textbf{1.04} $\pm$ 0.20 &    \textbf{0.67} $\pm$ 0.02 &    \textbf{2.73} $\pm$ 0.37 \\
167120  &            -0.05 $\pm$ 0.13 &   \textbf{0.38} $\pm$ 0.21 &            -0.45 $\pm$ 0.21 &    \textbf{0.66} $\pm$ 0.39 \\
168911  &   \textbf{19.55} $\pm$ 1.75 &   \textbf{3.85} $\pm$ 2.27 &            -1.62 $\pm$ 1.77 &            -0.44 $\pm$ 2.24 \\
168912  &   \textbf{25.96} $\pm$ 6.75 &   \textbf{4.63} $\pm$ 4.89 &    \textbf{8.83} $\pm$ 0.57 &    \textbf{9.50} $\pm$ 8.18 \\
3       &  \textbf{69.43} $\pm$ 10.67 &  \textbf{67.37} $\pm$ 0.49 &   \textbf{19.37} $\pm$ 2.69 &  \textbf{49.68} $\pm$ 17.90 \\
31      &   \textbf{17.25} $\pm$ 1.87 &  \textbf{10.56} $\pm$ 2.73 &    \textbf{8.29} $\pm$ 2.98 &    \textbf{9.46} $\pm$ 2.93 \\
3917    &    \textbf{0.92} $\pm$ 1.23 &           -0.23 $\pm$ 1.24 &    \textbf{1.36} $\pm$ 0.23 &            -1.40 $\pm$ 0.23 \\
53      &  \textbf{17.78} $\pm$ 14.46 &  \textbf{18.06} $\pm$ 3.86 &   \textbf{20.00} $\pm$ 1.81 &  \textbf{31.85} $\pm$ 13.24 \\
7592    &    \textbf{7.64} $\pm$ 2.28 &   \textbf{3.89} $\pm$ 3.50 &    \textbf{4.28} $\pm$ 0.08 &            -1.36 $\pm$ 3.21 \\
9952    &    \textbf{2.13} $\pm$ 1.08 &   \textbf{2.98} $\pm$ 0.61 &            -0.35 $\pm$ 0.63 &            -3.18 $\pm$ 0.31 \\
9977    &   \textbf{24.91} $\pm$ 9.65 &  \textbf{20.36} $\pm$ 8.15 &   \textbf{25.20} $\pm$ 0.29 &   \textbf{7.44} $\pm$ 11.08 \\
9981    &   \textbf{83.08} $\pm$ 4.49 &  \textbf{30.00} $\pm$ 9.35 &  \textbf{55.00} $\pm$ 14.58 &  \textbf{65.71} $\pm$ 11.61 \\
\bottomrule
\end{tabular}
\label{table:lr}
\end{table}

%% file: tables/NNBenchmark_table.tex
\begin{table}
\centering
\caption{\textbf{Average percent improvement and standard error for the best observed loss over five trials of \HomOpt{} over the base methodology for each dataset and method in the MLP Benchmark. Bold values indicate where \HomOpt{} outperforms the base methodology.}}
\label{tab:mlp-benchmark}
\begin{tabular}{ccccc}
\toprule
\multirow{2}{*}{Dataset} & \multicolumn{4}{c}{Method}             \\
 & \textbf{bayes} &             \textbf{random} &              \textbf{smac} &                \textbf{tpe}                    \\
\midrule
10101   &   \textbf{3.72} $\pm$ 0.57 &   \textbf{0.98} $\pm$ 1.65 &   \textbf{7.14} $\pm$ 3.69 &   \textbf{3.90} $\pm$ 1.65 \\
146818  &  \textbf{57.50} $\pm$ 3.64 &  \textbf{17.50} $\pm$ 3.06 &          -32.00 $\pm$ 8.00 &  \textbf{20.00} $\pm$ 7.28 \\
146822  &  \textbf{59.22} $\pm$ 0.73 &  \textbf{43.59} $\pm$ 4.51 &  \textbf{30.67} $\pm$ 0.67 &  \textbf{50.73} $\pm$ 1.42 \\
31      &  \textbf{40.00} $\pm$ 2.07 &  \textbf{35.56} $\pm$ 5.05 &  \textbf{25.83} $\pm$ 4.04 &   \textbf{4.44} $\pm$ 4.44 \\
53      &  \textbf{78.46} $\pm$ 5.10 &  \textbf{44.62} $\pm$ 2.88 &  \textbf{40.00} $\pm$ 2.23 &  \textbf{40.00} $\pm$ 8.37 \\
\bottomrule
\end{tabular}
\label{table:nn}
\end{table}

%% file: tables/XGBoostBenchmark_table.tex
\begin{table}
\centering
\caption{\textbf{Average percent improvement and standard error for the best observed loss over five trials of \HomOpt{} over the base methodology for each dataset and method in the XGBoost Benchmark. Bold values indicate where \HomOpt{} outperforms the base methodology.}}
\label{tab:xgboost-benchmark}
\begin{tabular}{ccccc}
\toprule
\multirow{2}{*}{Dataset} & \multicolumn{4}{c}{Method}             \\
 & \textbf{bayes} &             \textbf{random} &              \textbf{smac} &                \textbf{tpe}                    \\
\midrule
10101   &  \textbf{17.33} $\pm$ 2.21 &  \textbf{13.79} $\pm$ 1.54 &   -10.91 $\pm$ 1.82 &           -5.45 $\pm$ 0.91 \\
146822  &          -11.43 $\pm$ 5.35 &          -22.86 $\pm$ 3.50 &   -17.14 $\pm$ 5.35 &           -5.71 $\pm$ 5.71 \\
168912  &           -2.67 $\pm$ 6.18 &           -9.33 $\pm$ 6.53 &   -11.43 $\pm$ 5.35 &   \textbf{1.25} $\pm$ 5.00 \\
31      &           -5.00 $\pm$ 4.59 &   \textbf{6.67} $\pm$ 2.08 &    -7.50 $\pm$ 1.25 &  \textbf{10.00} $\pm$ 2.08 \\
3917    &          -27.00 $\pm$ 2.00 &   \textbf{9.60} $\pm$ 4.66 &    -8.18 $\pm$ 4.64 &           -3.48 $\pm$ 5.74 \\
53      &   \textbf{5.45} $\pm$ 6.17 &           -8.00 $\pm$ 3.74 &    -4.00 $\pm$ 8.12 &   \textbf{0.00} $\pm$ 5.48 \\
9952    &           -1.28 $\pm$ 2.48 &   \textbf{6.27} $\pm$ 1.90 &    -1.28 $\pm$ 1.28 &   \textbf{0.43} $\pm$ 3.10 \\
\bottomrule
\end{tabular}
\label{table:xgboost}
\end{table}

%% file: tables/RandomForestBenchmark_test_loss_table.tex
\begin{table}
\centering
\caption{\textbf{Average percent improvement and standard error for the corresponding test loss (1-accuracy) at best observed loss over five trials of \HomOpt{} over the base methodology for each dataset and method in the RandomForestBenchmark. Bold values indicate where \HomOpt{} outperforms the base methodology.}}
\label{tab:RandomForestBenchmark_test_loss}
\begin{tabular}{ccccc}
\toprule
method &              \textbf{bayes} &            \textbf{random} &               \textbf{smac} &               \textbf{tpe} \\
dataset &                             &                            &                             &                            \\
\midrule
10101   &           -11.76 $\pm$ 3.22 &           -2.22 $\pm$ 5.98 &           -13.33 $\pm$ 2.22 &           -9.47 $\pm$ 5.10 \\
12      &   \textbf{11.11} $\pm$ 6.09 &   \textbf{2.86} $\pm$ 5.35 &           -11.43 $\pm$ 5.35 &          -36.00 $\pm$ 9.80 \\
146818  &            -8.57 $\pm$ 5.71 &  \textbf{20.00} $\pm$ 8.89 &           -0.00 $\pm$ 10.46 &   \textbf{7.50} $\pm$ 9.35 \\
146821  &   \textbf{56.92} $\pm$ 5.76 &  \textbf{4.00} $\pm$ 16.00 &          -26.67 $\pm$ 26.67 &         -86.67 $\pm$ 34.32 \\
146822  &           -15.29 $\pm$ 4.40 &          -32.86 $\pm$ 8.63 &    \textbf{5.88} $\pm$ 6.17 &          -35.71 $\pm$ 3.91 \\
167119  &           -16.40 $\pm$ 1.16 &           -7.15 $\pm$ 0.82 &    \textbf{0.27} $\pm$ 0.85 &           -0.94 $\pm$ 0.89 \\
168911  &    \textbf{2.76} $\pm$ 2.01 &           -5.00 $\pm$ 1.91 &            -3.21 $\pm$ 1.54 &   \textbf{2.86} $\pm$ 1.34 \\
168912  &            -0.54 $\pm$ 3.35 &   \textbf{2.11} $\pm$ 3.16 &            -1.11 $\pm$ 4.94 &           -1.11 $\pm$ 1.67 \\
31      &   \textbf{15.17} $\pm$ 5.52 &  \textbf{15.20} $\pm$ 5.28 &  \textbf{12.80} $\pm$ 10.23 &  \textbf{15.56} $\pm$ 3.95 \\
3917    &    \textbf{2.76} $\pm$ 4.28 &   \textbf{8.97} $\pm$ 3.71 &    \textbf{5.33} $\pm$ 3.27 &           -3.08 $\pm$ 4.93 \\
53      &   \textbf{10.43} $\pm$ 3.53 &   \textbf{8.70} $\pm$ 5.99 &   \textbf{11.82} $\pm$ 1.82 &           -6.32 $\pm$ 6.09 \\
9952    &   \textbf{22.00} $\pm$ 4.25 &           -6.27 $\pm$ 2.93 &    \textbf{3.16} $\pm$ 4.35 &  \textbf{16.77} $\pm$ 2.42 \\
9981    &  \textbf{23.33} $\pm$ 12.47 &   \textbf{3.33} $\pm$ 9.72 &         -120.00 $\pm$ 37.42 &   \textbf{4.00} $\pm$ 9.80 \\
\bottomrule
\end{tabular}
\end{table}

%% file: tables/SVMBenchmark_test_loss_table.tex
\begin{table}
\centering
\caption{\textbf{Average percent improvement and standard error for the corresponding test loss (1-accuracy) at best observed loss over five trials of \HomOpt{} over the base methodology for each dataset and method in the SVMBenchmark. Bold values indicate where \HomOpt{} outperforms the base methodology.}}
\label{tab:SVMBenchmark_test_loss}
\begin{tabular}{ccccc}
\toprule
method &              \textbf{bayes} &             \textbf{random} &               \textbf{smac} &                \textbf{tpe} \\
dataset &                             &                             &                             &                             \\
\midrule
10101   &            -3.75 $\pm$ 2.50 &            -5.00 $\pm$ 2.34 &    \textbf{2.35} $\pm$ 2.35 &            -5.00 $\pm$ 2.34 \\
12      &    \textbf{0.22} $\pm$ 0.14 &    \textbf{0.22} $\pm$ 0.14 &    \textbf{4.11} $\pm$ 3.84 &    \textbf{1.44} $\pm$ 1.31 \\
146818  &  \textbf{32.26} $\pm$ 19.75 &  \textbf{62.58} $\pm$ 15.72 &  \textbf{64.52} $\pm$ 16.42 &         -105.00 $\pm$ 74.96 \\
146821  &    \textbf{0.00} $\pm$ 0.00 &  \textbf{27.33} $\pm$ 41.12 &  \textbf{59.23} $\pm$ 24.18 &  \textbf{18.46} $\pm$ 18.46 \\
146822  &   \textbf{57.08} $\pm$ 1.06 &            -5.71 $\pm$ 5.91 &   \textbf{17.50} $\pm$ 5.17 &   \textbf{53.06} $\pm$ 2.41 \\
168911  &   \textbf{33.15} $\pm$ 8.37 &          -71.51 $\pm$ 20.25 &           -9.64 $\pm$ 16.05 &  \textbf{25.37} $\pm$ 11.27 \\
168912  &  \textbf{36.48} $\pm$ 15.05 &  \textbf{35.40} $\pm$ 16.11 &  \textbf{37.20} $\pm$ 15.49 &  \textbf{23.12} $\pm$ 14.28 \\
3       &  \textbf{56.34} $\pm$ 23.00 &          -31.72 $\pm$ 99.04 &   \textbf{94.25} $\pm$ 0.48 &       -1224.00 $\pm$ 708.84 \\
31      &    \textbf{0.00} $\pm$ 0.00 &    \textbf{0.00} $\pm$ 0.00 &    \textbf{0.00} $\pm$ 0.00 &    \textbf{0.00} $\pm$ 0.00 \\
3917    &            -0.00 $\pm$ 0.96 &            -0.00 $\pm$ 0.96 &            -0.00 $\pm$ 0.96 &            -0.00 $\pm$ 0.96 \\
53      &  \textbf{37.46} $\pm$ 14.15 &           -2.86 $\pm$ 24.42 &          -23.33 $\pm$ 22.92 &          -62.22 $\pm$ 35.94 \\
9952    &          -16.94 $\pm$ 22.46 &           -15.25 $\pm$ 6.72 &    \textbf{9.41} $\pm$ 2.96 &    \textbf{0.60} $\pm$ 8.32 \\
9981    &            -0.65 $\pm$ 0.43 &            -1.09 $\pm$ 0.69 &            -1.54 $\pm$ 0.56 &            -0.65 $\pm$ 0.55 \\
\bottomrule
\end{tabular}
\end{table}

%% file: tables/LRBenchmark_test_loss_table.tex
\begin{table}
\centering
\caption{\textbf{Average percent improvement and standard error for the corresponding test loss (1-accuracy) at best observed loss over five trials of \HomOpt{} over the base methodology for each dataset and method in the LRBenchmark. Bold values indicate where \HomOpt{} outperforms the base methodology.}}
\label{tab:LRBenchmark_test_loss}
\begin{tabular}{ccccc}
\toprule
method &              \textbf{bayes} &            \textbf{random} &               \textbf{smac} &                \textbf{tpe} \\
dataset &                             &                            &                             &                             \\
\midrule
10101   &            -4.44 $\pm$ 1.11 &           -6.67 $\pm$ 2.08 &    \textbf{5.26} $\pm$ 2.35 &    \textbf{8.00} $\pm$ 2.55 \\
12      &  \textbf{64.71} $\pm$ 11.91 &  \textbf{36.00} $\pm$ 7.48 &   \textbf{43.33} $\pm$ 6.67 &  \textbf{52.50} $\pm$ 17.74 \\
146212  &  \textbf{39.01} $\pm$ 15.89 &  \textbf{45.74} $\pm$ 7.84 &   \textbf{36.41} $\pm$ 5.63 &   \textbf{44.18} $\pm$ 7.26 \\
146606  &   \textbf{16.73} $\pm$ 1.74 &   \textbf{3.42} $\pm$ 1.37 &    \textbf{5.84} $\pm$ 0.12 &    \textbf{5.89} $\pm$ 1.97 \\
146818  &           -8.57 $\pm$ 16.66 &  \textbf{20.00} $\pm$ 7.07 &  \textbf{11.11} $\pm$ 11.65 &          -33.33 $\pm$ 10.54 \\
146821  &   \textbf{30.34} $\pm$ 4.55 &  \textbf{29.66} $\pm$ 3.55 &   \textbf{18.46} $\pm$ 5.63 &            -4.55 $\pm$ 3.80 \\
146822  &   \textbf{22.45} $\pm$ 8.09 &           -2.86 $\pm$ 9.52 &           -16.30 $\pm$ 1.89 &   \textbf{36.73} $\pm$ 2.66 \\
14965   &    \textbf{2.85} $\pm$ 1.97 &           -0.99 $\pm$ 1.64 &    \textbf{1.65} $\pm$ 2.42 &    \textbf{1.03} $\pm$ 2.29 \\
167119  &    \textbf{4.02} $\pm$ 0.34 &           -0.25 $\pm$ 0.43 &    \textbf{1.17} $\pm$ 0.19 &    \textbf{3.58} $\pm$ 0.43 \\
167120  &    \textbf{0.30} $\pm$ 0.20 &   \textbf{0.96} $\pm$ 0.14 &    \textbf{0.58} $\pm$ 0.48 &    \textbf{1.16} $\pm$ 0.62 \\
168911  &    \textbf{4.51} $\pm$ 1.29 &           -8.57 $\pm$ 1.19 &            -8.20 $\pm$ 1.16 &            -2.46 $\pm$ 2.73 \\
168912  &   \textbf{20.79} $\pm$ 5.17 &   \textbf{1.11} $\pm$ 4.25 &    \textbf{1.13} $\pm$ 0.96 &   \textbf{13.82} $\pm$ 5.36 \\
3       &  \textbf{67.89} $\pm$ 12.47 &  \textbf{57.78} $\pm$ 3.23 &   \textbf{33.33} $\pm$ 4.22 &  \textbf{46.15} $\pm$ 18.84 \\
31      &   \textbf{10.00} $\pm$ 6.43 &  \textbf{10.71} $\pm$ 7.41 &   \textbf{12.86} $\pm$ 4.16 &   \textbf{19.37} $\pm$ 4.98 \\
3917    &    \textbf{1.88} $\pm$ 2.90 &          -14.29 $\pm$ 2.53 &    \textbf{7.27} $\pm$ 0.74 &    \textbf{1.88} $\pm$ 2.90 \\
53      &   \textbf{0.87} $\pm$ 17.25 &  \textbf{24.17} $\pm$ 6.37 &    \textbf{5.56} $\pm$ 3.04 &  \textbf{18.26} $\pm$ 12.33 \\
7592    &    \textbf{9.55} $\pm$ 2.40 &   \textbf{5.15} $\pm$ 4.00 &    \textbf{1.66} $\pm$ 0.28 &            -1.59 $\pm$ 3.88 \\
9952    &    \textbf{0.32} $\pm$ 1.50 &           -0.16 $\pm$ 0.60 &            -2.58 $\pm$ 0.16 &            -2.15 $\pm$ 1.00 \\
9977    &   \textbf{20.85} $\pm$ 8.66 &  \textbf{21.50} $\pm$ 7.28 &   \textbf{18.12} $\pm$ 0.63 &    \textbf{7.77} $\pm$ 9.79 \\
9981    &           -3.33 $\pm$ 20.68 &         -30.00 $\pm$ 14.58 &  \textbf{10.00} $\pm$ 11.30 &   \textbf{5.71} $\pm$ 16.66 \\
\bottomrule
\end{tabular}
\end{table}

%% file: tables/NNBenchmark_test_loss_table.tex
\begin{table}
\centering
\caption{\textbf{Average percent improvement and standard error for the corresponding test loss (1-accuracy) at best observed loss over five trials of \HomOpt{} over the base methodology for each dataset and method in the NNBenchmark. Bold values indicate where \HomOpt{} outperforms the base methodology.}}
\label{tab:NNBenchmark_test_loss}
\begin{tabular}{ccccc}
\toprule
method &             \textbf{bayes} &             \textbf{random} &              \textbf{smac} &               \textbf{tpe} \\
dataset &                            &                             &                            &                            \\
\midrule
10101   &   \textbf{0.00} $\pm$ 1.76 &    \textbf{1.05} $\pm$ 1.97 &           -0.00 $\pm$ 2.88 &           -3.33 $\pm$ 2.22 \\
146818  &         -57.14 $\pm$ 10.10 &          -25.00 $\pm$ 11.18 &   \textbf{1.82} $\pm$ 6.68 &          -32.50 $\pm$ 9.35 \\
146822  &  \textbf{39.13} $\pm$ 4.96 &  \textbf{20.00} $\pm$ 11.42 &  \textbf{12.00} $\pm$ 4.90 &  \textbf{30.59} $\pm$ 6.81 \\
31      &   \textbf{3.85} $\pm$ 2.43 &    \textbf{8.46} $\pm$ 3.73 &           -3.48 $\pm$ 1.63 &   \textbf{9.23} $\pm$ 4.65 \\
53      &  \textbf{21.54} $\pm$ 7.46 &   \textbf{21.33} $\pm$ 3.27 &  \textbf{20.00} $\pm$ 4.71 &   \textbf{1.54} $\pm$ 9.23 \\
\bottomrule
\end{tabular}
\end{table}

%% file: tables/XGBoostBenchmark_test_loss_table.tex
\begin{table}
\centering
\caption{\textbf{Average percent improvement and standard error for the corresponding test loss (1-accuracy) at best observed loss over five trials of \HomOpt{} over the base methodology for each dataset and method in the XGBoostBenchmark. Bold values indicate where \HomOpt{} outperforms the base methodology.}}
\label{tab:XGBoostBenchmark_test_loss}
\begin{tabular}{ccccc}
\toprule
method &             \textbf{bayes} &            \textbf{random} &             \textbf{smac} &               \textbf{tpe} \\
dataset &                            &                            &                           &                            \\
\midrule
10101   &          -30.67 $\pm$ 5.42 &           -8.89 $\pm$ 4.16 &         -13.33 $\pm$ 4.16 &   \textbf{7.62} $\pm$ 6.67 \\
146818  &  \textbf{13.33} $\pm$ 8.89 &         -11.43 $\pm$ 12.29 &         -20.00 $\pm$ 7.28 &         -64.00 $\pm$ 21.35 \\
146822  &           -5.33 $\pm$ 3.89 &           -2.67 $\pm$ 4.52 &         -15.71 $\pm$ 6.93 &   \textbf{2.35} $\pm$ 6.06 \\
168912  &   \textbf{1.62} $\pm$ 2.51 &          -10.29 $\pm$ 1.46 &  \textbf{8.72} $\pm$ 2.08 &           -3.53 $\pm$ 4.78 \\
31      &  \textbf{11.67} $\pm$ 6.77 &           -2.50 $\pm$ 4.08 &         -17.14 $\pm$ 2.43 &  \textbf{15.71} $\pm$ 2.90 \\
3917    &          -32.50 $\pm$ 2.43 &           -2.00 $\pm$ 5.44 &         -33.33 $\pm$ 2.95 &          -27.50 $\pm$ 8.50 \\
53      &          -15.29 $\pm$ 8.44 &          -12.63 $\pm$ 6.14 &         -10.00 $\pm$ 3.24 &           -7.78 $\pm$ 2.83 \\
9952    &   \textbf{3.51} $\pm$ 3.80 &  \textbf{10.82} $\pm$ 1.23 &  \textbf{4.56} $\pm$ 1.43 &           -4.29 $\pm$ 1.07 \\
\bottomrule
\end{tabular}
\end{table}